%% file: main.tex
\documentclass[conference]{IEEEtran}
\IEEEoverridecommandlockouts

\def\forsubmission{}

\input{include/packages}
\input{include/cmds}

\input{include/shorthand}
\usepackage{placeins}
\def\FIGDIR{./figures}

\begin{document}

\title{Characterizing the Accuracy-Efficiency Trade-off of Low-rank Decomposition in Language Models}

\author{\IEEEauthorblockN{Chakshu Moar}
\IEEEauthorblockA{\textit{Electrical Engineering and Computer Science} \\
\textit{University of California, Irvine}\\
Irvine CA, USA \\
cmoar@uci.edu}
\and
\IEEEauthorblockN{Faraz Tahmasebi}
\IEEEauthorblockA{\textit{Electrical Engineering and Computer Science} \\
\textit{University of California, Irvine}\\
Irvine CA, USA \\
tahmasef@uci.edu}
\and
\IEEEauthorblockN{Michael Pellauer}
\IEEEauthorblockA{\textit{NVIDIA}\\
Westford MA, USA \\
mpellauer@nvidia.com}
\and
\IEEEauthorblockN{Hyoukjun Kwon}
\IEEEauthorblockA{\textit{Electrical Engineering and Computer Science} \\
\textit{University of California, Irvine}\\
Irvine CA, USA \\
hyoukjun.kwon@uci.edu}
}


\maketitle

\input{sections/00_Abstract}
\input{sections/01_Introduction}
\input{sections/02_background}
\input{sections/03_Methodology}
\input{sections/04_CaseStudy}
\input{sections/05_RelatedWorks}
\input{sections/06_Conclusion}
\input{sections/07_Acknowledgement}


\bibliography{ref}
\bibliographystyle{ieeetr}

\end{document}

%% file: include/packages.tex
\usepackage{cite}
\usepackage{amsmath,amssymb,amsfonts,amsthm}
\usepackage{graphicx}
\usepackage[dvipsnames]{xcolor}
\usepackage[final]{microtype}
\usepackage[italic]{mathastext}
\usepackage{libertine}
\usepackage[T1]{fontenc}
\usepackage{textcomp}
\usepackage[varqu,varl]{zi4}
\usepackage[all]{nowidow}
\usepackage[keeplastbox]{flushend}
\usepackage{fancyhdr}

\usepackage{makecell}

\usepackage{algorithmic}
\theoremstyle{definition}
\newtheorem{definition}{Def.}
\newtheorem{theorem}{Theorem}
\newtheorem{proposition}{Prop.}

\usepackage{newfloat}

\usepackage[small,compact]{titlesec}
\usepackage[linesnumbered]{algorithm2e}
\usepackage{caption}
\usepackage{color}
\usepackage[normalem]{ulem}
\usepackage{xspace}
\usepackage{booktabs} 

\usepackage{makecell}
\usepackage{multirow}
\usepackage{comment}
\usepackage{hhline}
\usepackage[para]{threeparttable} 
\makeatletter
\def\TPT@doparanotes{\par
   \prevdepth\z@ \TPT@hsize
   \TPTnoteSettings
   \parindent\z@ \pretolerance 8
   \linepenalty 200
   \renewcommand\item[1][]{\relax\ifhmode \begingroup
       \unskip
       \advance\hsize 10em 
       \penalty -45 \hskip\z@\@plus\hsize \penalty-19
       \hskip .07\hsize \penalty 9999 \hskip-.15\hsize
       \hskip .01\hsize\@plus-\hsize\@minus.01\hsize 
       \hskip 6em\@plus .3em
      \endgroup\fi
      \tnote{##1}\,\ignorespaces}%
   \let\TPToverlap\relax
   \def\endtablenotes{\par}%
}
\makeatother

\usepackage{hyperref}

\RequirePackage[nameinlink]{cleveref} 
\crefname{chapter}{Chapter}{Chapters}
\crefname{section}{Section}{Sections}
\crefname{subsection}{Subsection}{Subsections}
\crefname{equation}{Equation}{Equations}
\crefname{definition}{Definition}{Definitions}
\crefname{assumption}{Assumption}{Assumptions}
\crefname{theorem}{Theorem}{Theorems}
\crefname{figure}{Figure}{Figures}
\crefname{table}{Table}{Tables}
\crefname{BOX}{Box}{Boxes}
\let\autoref\cref 

\DeclareMathOperator*{\argmin}{arg\,min}

%% file: include/cmds.tex
\newcommand{\insertFigure}[2]{
    \begin{figure}[t]
        \centering
        \includegraphics[width=\linewidth]{\FIGDIR/#1.pdf}
        \vspace{-4mm}
        \caption{\small #2}
        \vspace{-2mm}
        \label{fig:#1}
    \end{figure}
}

\newcommand{\insertWideFigure}[2]{
    \begin{figure*}[h]
        \centering
        \includegraphics[width=\textwidth]{\FIGDIR/#1.pdf}
        \vspace{-4mm}
        \caption{\small #2}
        \vspace{-2mm}
        \label{fig:#1}
    \end{figure*}
}

\DeclareFloatingEnvironment[
  fileext=lob,
  listname={List of Boxes},
  name=Box,
  placement=htp,
]{BOX}

\ifx\forsubmission\undefined
\newcommand{\TODO}[1]{\textcolor{red}{TODO: #1}}
\newcommand{\CM}[1]{\textcolor{magenta}{CM: #1}}
\newcommand{\FT}[1]{\textcolor{orange}{FT: #1}}
\newcommand{\HK}[1]{\textcolor{blue}{HK: #1}}
\newcommand{\MP}[1]{\textcolor{teal}{MP: #1}}
\newcommand{\revision}[1]{\textcolor{blue}{#1}}
\else
\newcommand{\TODO}[1]{\textcolor{red}{}}
\newcommand{\CM}[1]{\textcolor{magenta}{}}
\newcommand{\FT}[1]{\textcolor{orange}{}}
\newcommand{\HK}[1]{\textcolor{blue}{}}
\newcommand{\MP}[1]{\textcolor{teal}{}}
\newcommand{\revision}[1]{#1}
\fi

\newcommand{\squishlist}{
 \begin{list}{$\bullet$}
  { \setlength{\itemsep}{0pt}
     \setlength{\parsep}{3pt}
     \setlength{\topsep}{3pt}
     \setlength{\partopsep}{0pt}
     \setlength{\leftmargin}{1.5em}
     \setlength{\labelwidth}{1em}
     \setlength{\labelsep}{0.5em} } }

\newcommand{\squishlisttwo}{
 \begin{list}{$\bullet$}
  { \setlength{\itemsep}{0pt}
     \setlength{\parsep}{0pt}
    \setlength{\topsep}{0pt}
    \setlength{\partopsep}{0pt}
    \setlength{\leftmargin}{2em}
    \setlength{\labelwidth}{1.5em}
    \setlength{\labelsep}{0.5em} } }

\newcommand{\squishend}{
  \end{list}  }

\newcommand{\betterparagraph}[1]{\noindent \textbf{#1. }}

\newcommand{\norm}[1]{\left\lVert#1\right\rVert}

\crefname{algorithm}{Algorithm}{Algorithms}

%% file: sections/00_Abstract.tex
\begin{abstract}
Recent large language models (LLMs) employ billions of parameters to enable broad problem-solving capabilities.
Such language models also tend to be memory-bound because of the dominance of matrix-vector and matrix-matrix multiplications with low arithmetic intensity.
%
%
Therefore, optimizing the memory footprint and traffic is an important optimization direction for LLMs today.
Model compression methods such as quantization and parameter pruning have been actively explored to achieve memory footprint and traffic optimization.
However, the accuracy-efficiency trade-off of rank pruning (i.e., low-rank decomposition) for LLMs is not well-understood yet.
Therefore, in this work, we characterize the accuracy-efficiency trade-off of a low-rank decomposition method, specifically Tucker decomposition, on recent language models, including an open-source LLM, Llama 2.

We formalize the low-rank decomposition design space and show that the decomposition design space is enormous (e.g., O($2^{39}$) for Llama2-7B).
To navigate such a vast design space, we formulate it and perform thorough case studies of accuracy-efficiency trade-offs using six widely used LLM benchmarks on BERT and Llama 2 models.
Our results show that we can achieve a 9\% model size reduction with minimal accuracy drops, which range from 4\%p (\%p refers to "percentage point," which refers to the absolute difference between two percentage numbers; 74\% -> 78\% = 4\%p increase) to 10\%p, depending on the difficulty of the benchmark, without any retraining to recover accuracy after decomposition.
The results show that low-rank decomposition can be a promising direction for LLM-based applications that require real-time service at scale (e.g., AI agent and real-time coding assistant), where the latency is as important as the model accuracy.

\end{abstract}

%% file: sections/01_Introduction.tex
\section{Introduction}
\label{sec:intro}

Large language models (LLMs) such as GPT-4~\cite{openai2023gpt4} have opened a new era of artificial intelligence (AI) technologies based on broad problem-solving capabilities and even encompassing generative tasks~\cite{huang2023realtime, ni2023lever} interfaced with natural languages.
%
%
This success was mainly driven by a massive amount of training data~\cite{openai2023gpt4}, leading to many model parameters to learn effectively.
The number of parameters in the state-of-the-art models reaches up to 70 billion parameters on a popular open LLM, Llama 2~\cite{touvron2023llama}, which translates to 140 GB of memory in FP16 data format just for model parameters (i.e., weights).
Such a large memory requirement is beyond typical on-board memory sizes in a single GPU (e.g., 80GB in NVIDIA A100 and H100), which presents a major challenge for providing services like ChatGPT at scale.

Although LLM variants in smaller scales (e.g., Llama2-7B~\cite{touvron2023llama}) exist, their memory requirements are still high compared to those of convolutional neural networks (CNNs).
For example, Llama 2-7B has 268.5$\times$ more parameters compared to ResNet50~\cite{he2016deep}.
As an additional challenge, these rising footprints have been paired with decreased data reuse compared to CNNs.
This is because state-of-the-art LLMs are based on the Transformer~\cite{vaswani2017attention} architecture, and its operators have a significantly low compute-to-model size ratio, as summarized in~\autoref{tab:arithmetic_intensity}.
This low compute-to-model size ratio and large memory footprint indicate that optimizations for LLM inferences need to focus on the memory side rather than increasing peak throughput.

\insertFigure{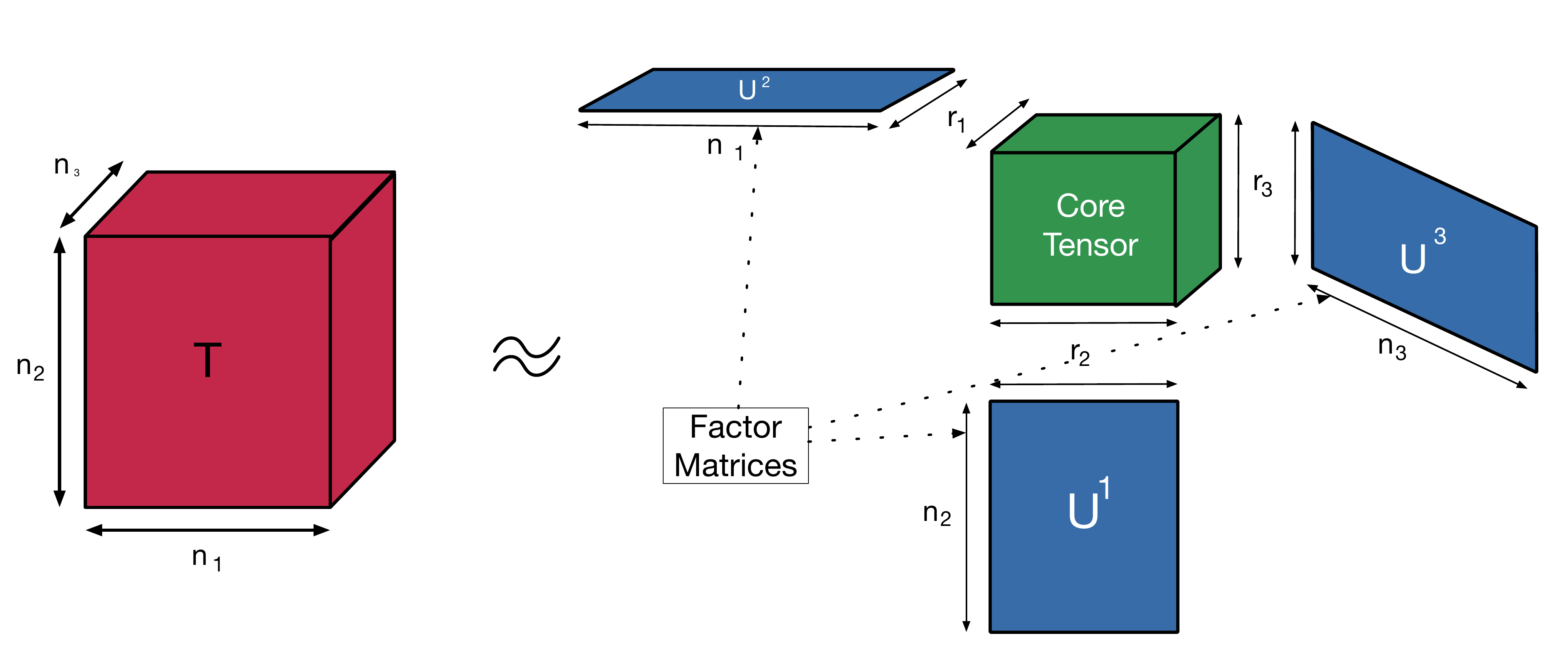}{An illustration of Tucker Decomposition. A three-dimensional tensor T can be decomposed into one core tensor and three-factor matrices, $U^{1}$, $U^{2}$, and $U^{3}$. the dimension of the core tensor corresponds to the rank of the decomposition.}

To address the challenge, the research community has been actively exploring model compression methodologies (e.g., sparsity~\cite{frantar2023sparsegpt} and quantization~\cite{kim2021bert, yao2022zeroquant,dettmers2022gpt3}).
Among them, low-rank decomposition is an approach that analyzes the dimensionality of a tensor and prunes minor components  (i.e., ranks) in the decomposed dimensions.
One of the low-rank decomposition methodologies, Tucker decomposition is the generalization of principal component analysis (PCA) for high-dimensional tensors~\cite{zare2018extension}.
As illustrated in~\autoref{fig:Tucker_Decomposition}, the low-rank Tucker decomposition method decomposes a tensor into a series of tensor contractions (or matrix multiplications if the base tensors are two-dimensional).
When performing the conversion, we prune the rank of the decomposed tensors by removing unimportant dimensions, similar to the dimension reduction methods based on PCA.
The pruned ranks lead to smaller memory requirements compared to the non-decomposed tensor, but this also leads to approximate tensor reconstruction, which can degrade the task performance (e.g., accuracy and perplexity).
%

The application of low-rank decomposition has been actively explored in the computer vision domain~\cite{kong2017low, yu2017compressing, gusak2019automated, wang2023cuttlefish}.
Based on such successful application cases, researchers are exploring the case on large language models~\cite{xu2023tensorgpt}.
However, unlike weight pruning (sparsity) and quantization (data precision), the trade-offs among task performance (e.g., model accuracy), computational performance (e.g., latency), and energy efficiency of low-rank decomposition targeting language models are not well-understood yet. 
In addition, we demonstrate that low-rank decomposition has a large design space originating from many possible choices (e.g., the number of pruned ranks, the choice of decomposed layers and tensors, and so on), so understanding the trade-off space is difficult.
For example, when we apply Tucker decomposition on the Llama2-7B model, our design-space formulation (~\autoref{subsec:decomp_space_formulation}) reveals that there exist $\mathbf{O(2^{39})}$ possible ways of applying Tucker decomposition, even if we apply it to the same set of tensors in each layer with the same pruned ranks across all the layers.
%

Therefore, in this work, we first formalize the design space of the low-rank decomposition on recent language models based on Transformers, then characterize the trade-off space among task performance, computational performance, and energy efficiency.
We perform a thorough profiling of the latency, energy, accuracy, and memory usage for running Llama2 and BERT after applying Tucker decomposition on 4$\times$ NVIDIA A100 GPUs (PCIe) with 80GB of memory for each.
We measure the metrics on six broadly adopted benchmarks for LLMs: AI2 Reasoning Challenge (ARC)~\cite{clark2018think} easy and challenge, HellaSwag~\cite{zellers2019hellaswag}, Massive Multitask Language Understanding (MMLU)~\cite{hendryckstest2021}, TruthfulQA~\cite{lin-etal-2022-truthfulqa}, WinoGrande~\cite{10.1145/3474381}.
The benchmarks include a variety of tasks oriented for LLMs, such as reasoning, truthfulness check, sentence completion, and commonsense reasoning.
In case studies, we show that low-rank decomposition can reduce the model size by 9\% without losing considerable model accuracy (4.5\% points to 10\% points).
In addition to the model size reduction, we observe a considerable reduction in the end-to-end latency by 4\% and energy by 5\%.
As part of this contribution, we study how to apply low-rank decomposition to the language models, performing a thorough characterization of the design space varying the number of pruned ranks and the choice of decomposed tensors (target layers and weight tensors in each target layer).
Based on the profiling results, We show that we can aggressively reduce the rank to one without losing significant accuracy and should carefully select the decomposition location to minimize the accuracy degradation.
Finally, we present a trade-off study between the model size reduction and resulting metrics (latency, energy, etc.).
The results show that we can reduce the model size by 9\% without losing considerable accuracy (on average, 10\% loss of accuracy), which provides 4\% latency and 5\% energy savings.
Our case study results show that low-rank factorization is a promising option to enable low-cost LLM-based services, such as Virtual Agent~\cite{google_agent_assist}, real-time animation generation~\cite{huang2023realtime}, real-time coding assistant~\cite{github_copilot}, AI assistant~\cite{microsoft_ai_assistant}, and so on.

\input{tables/tab_arithmetic_intensity}

We summarize our contributions as follows:

\begin{itemize}
    {\item We thoroughly explore the accuracy/performance trade-off of low-rank decomposition on recent language models, including a large language model: Llama-2-7B. }
    {\item We demystify the design space of low-rank decomposition on large language models by formally defining its dimensions.}   
    {\item Beyond the simple performance analysis, we also profile energy consumption and show that low-rank decomposition is an effective approach to enhancing energy efficiency.}
    {\item We analyze the sensitivity of low-rank decomposition and provide insights on how to best apply low-rank decomposition to language models.}
\end{itemize}

%% file: tables/tab_arithmetic_intensity.tex
\begin{table*}[t]
\caption{The model size and the number of computations (multiply-and-accumulate; MAC) of Transformer-based language models and convolutional neural network-based computer vision models. Language model data are based on the batch size of 1 and the sequence length of 128.}
\label{tab:arithmetic_intensity}
\resizebox{1.0\textwidth}{!}{
\centering
\begin{tabular}{|c|c|c|c|c|} 
\hline
\textbf{Models} & \textbf{Model Type} & \textbf{Model Size (FP16)} & \textbf{\# Operations (MACs)} & \textbf{Compute-to-model size ratio} \\
\hline
ResNet50~\cite{he2016deep}
& Computer Vision
& 51.1 MB
& 8.21 B
& 160.7
\\
\hline
BERT-Base~\cite{kenton2019bert} 
& Language Model 
& 219.0 MB
& 11.2 B
& 51.1
\\
\hline
Llama2-7B~\cite{touvron2023llama}
& Large Language Model 
& 13.4 GB
& 850.0 B
& 63.4
\\
\hline
\end{tabular}
}
\end{table*}

%% file: sections/02_background.tex
\section{Background and Motivation}
\label{sec:background}


We discuss backgrounds on the low-rank tensor decomposition and motivation toward our approach.

\subsection{Computation in Language Models (LMs)}
\label{subsec:challenges_for_llms}

\insertFigure{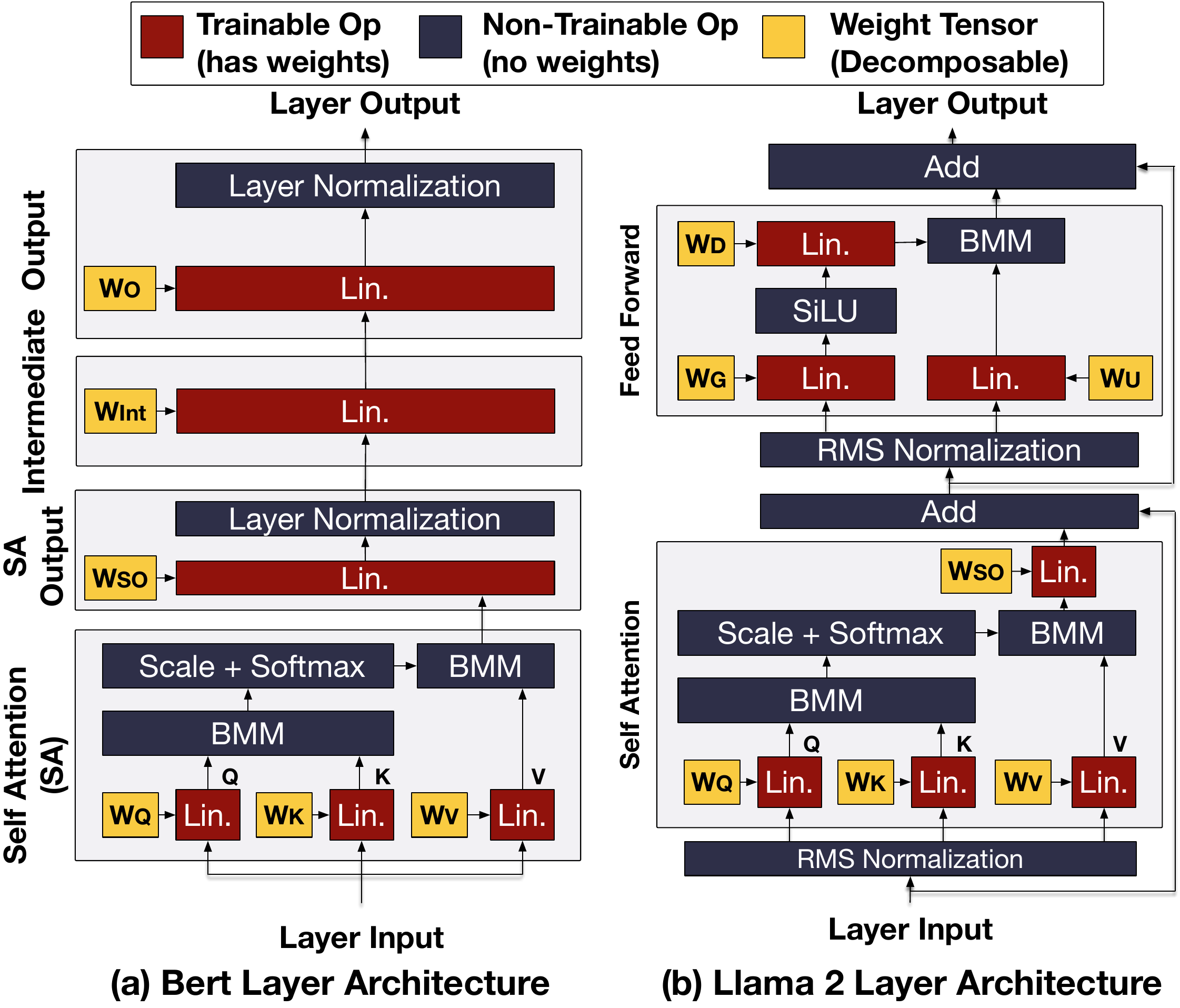}{The layer architecture of two recent language models: Bert~\cite{kenton2019bert} and Llama2~\cite{touvron2023llama}. Lin, BMM, and RMS refer to the linear layer, batched matrix multiplication, and root mean square, respectively. We highlight decomposable weight tensors
using yellow boxes.}

State-of-the-art language models~\cite{openai2023gpt4, touvron2023llama, anil2023gemini} today are mostly based on the Transformer~\cite{vaswani2017attention} architecture.
Although the original Transformer model was based on a full encoder-decoder structure, encoder-only~\cite{kenton2019bert} and decoder-only~\cite{openai2023gpt4} models emerged as effective solutions with high task performance (e.g., accuracy).
Such models include multiple layers connected in a linear architecture, as illustrated in~\autoref{fig:DesignSpace} (a).

Although the architecture of the layers can differ depending on the model, as~\autoref{fig:ModelArchitecture} shows, the majority of operators, such as linear layer and batched matrix multiplication (BMM), are common.
Among operators, linear operations (linear layer and BMM) account for approximately 75\% of the total execution time on CPUs~\cite{kim2023full}.
In addition, such linear operations involve a large number of parameters (e.g., 7 - 70 billion parameters for Llama2~\cite{touvron2023llama}), which results in large memory traffic and imposes a major challenge to efficient inference and generation.
A key concept that helps understand the large memory traffic challenge considering both computational throughput and memory bandwidth is the operational (or, arithmetic) intensity, which we discuss next.

\insertFigure{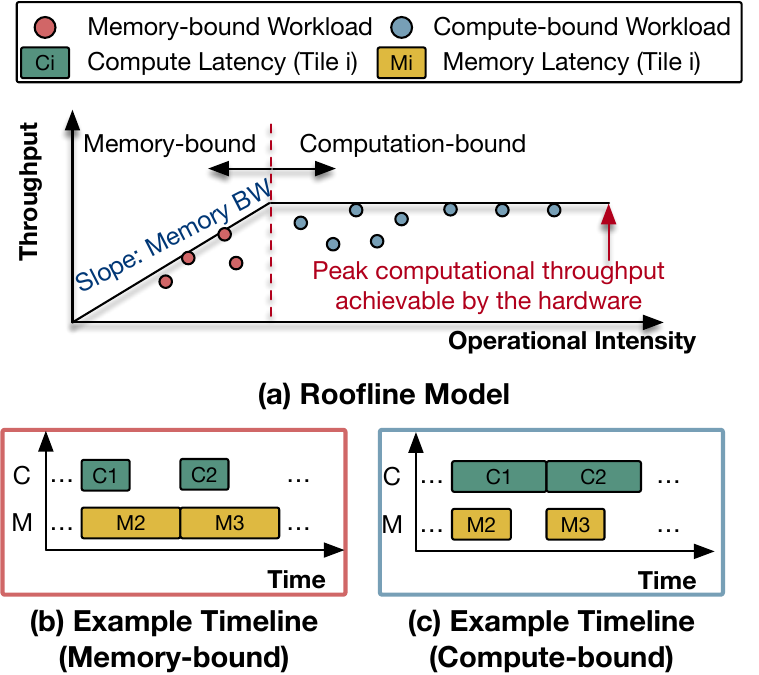}{A high-level example roofline that captures memory- and computation-boundness of workloads based on their operational intensity. BW, C, and M indicate bandwidth, compute, and memory, respectively. The example execution timelines in (b) and (c) show how the compute and memory latency overlaps for compute tiles for memory- and compute-bound scenarios.}

%

\subsection{Operational Intensity and Roofline Model}
\label{subsec:operational_intensity}
Operational Intensity (OI) refers to the number of operations per DRAM traffic bytes~\cite{williams2009roofline}, which can be represented as follows:

\begin{equation}
\label{eq:op_intensity}
OI = \frac{Num_{Ops}}{DRAM\_traffic}
\end{equation}

OI quantifies the compute-to-off-chip-communication ratio for running a workload on a computer system.
Workloads with low arithmetic intensity imply more number of DRAM accesses than those with high arithmetic intensity, which tends to shift the system bottleneck to the memory bandwidth from the computational throughput.
Such insights can be visually summarized in the roof-line model illustrated in~\autoref{fig:Roofline} (a).

The roofline model shows that the peak throughput of underlying hardware can be realized only if a workload has sufficiently large operational intensity.
If the operational intensity is low (memory-bound region in~\autoref{fig:Roofline}), the communication latency cannot be hidden within the computation latency when we apply computation-communication latency overlapping, as illustrated in~\autoref{fig:Roofline} (b).
That is, the communication latency becomes visible beyond the computation latency, which leads to lower throughput than the underlying hardware's peak throughput.
In such cases, as illustrated in the memory-bound region in~\autoref{fig:Roofline}, the overall throughput is constrained by the memory bandwidth.
%
%
Because operational intensity of a workload on hardware depends on software optimizations (e.g., if an inefficient compiler decides to flush data that will be reused soon after, the operational intensity may decrease), we use compute-to-model size for following discussions as the metric measuring the innate operational intensity characteristics of a workload.

\subsection{Challenges for Optimizing LM Performance}
\label{subsec:challenges_for_llms}

%
To understand the challenges for optimizing language model inference, we analyze the compute-to-model size ratios of BERT-Base~\cite{kenton2019bert} model with 110 million parameters fine-tuned on SQuAD dataset~\cite{rajpurkar-etal-2016-squad}, and Llama-2-7B~\cite{touvron2023llama} model with 7 billion parameters.
We compare the analysis results against a widely adopted convolutional neural network (CNN) model, Resnet50~\cite{he2016deep} targeting classification tasks on ImageNet~\cite{deng2009imagenet} dataset.

~\autoref{tab:arithmetic_intensity} shows the analysis results assuming FP16 precision.
The analysis reveals two major challenges for language models compared to the CNN: (1) large scale in model size and number of operations and (2) low compute-to-model size ratio.
Firstly, we observe that the model sizes and the number of operations of two language models are 4.3$\times-$262.3$\times$ and 1.4$\times-$103.5$\times$ larger than ResNet50, respectively.
Such a large scale of language models motivates optimizations for computational performance.

However, we also observe a major roadblock that hinders computational performance optimization: low compute-to-model size ratio of language models, which is 35.4\% lower than ResNet50, on average.
Accordingly, previous works reported that language model inference worklaods are memory-bound \cite{kao2023flat, kim2023full}.
In decoder-only models popular today~\cite{touvron2023llama, openai2023gpt4}, such a characteristic originates from the heavy use of matrix-vector multiplication operations, which are known to have even lower arithmetic intensity than matrix-matrix multiplication\cite{kim2023full}.

Our analysis indicates that the correct approach to optimize the computational performance of language models needs to focus on reducing memory bandwidth requirement, not on-chip computational latency.
As a such optimization method, we explore low-rank decomposition of parameter tensors, which shifts the memory bandwidth/size overhead toward the computational overhead.
We discuss low-rank decomposition in the LM context next.

\subsection{Tucker Decomposition (TKD) for Language Models}
\label{subsec:bg_tucker_decomposition}

Low-rank decomposition (LD) refers to techniques that factorize tensors into multiple smaller tensors, matrices, or vectors and prune unimportant factors in the decomposed space.
That is, LD can be utilized for compressing LMs with proper choice of the pruning factor. 
Popular LD approaches include Canonical Polyadic Decomposition (CPD)\cite{Harshman1970FoundationsOT}, Tensor-Train Decomposition (TTD) \cite{GNNTensorTrain, nips2020tensortrain}, and Tucker Decomposition (TKD) \cite{D-Tucker}.

\betterparagraph{Motivation toward TKD} 
%
CPD can have unstable convergence~\cite{doi:10.1137/06066518X, Lebedev2014SpeedingupCN}, which imposes a major challenge to the model training process.
Compared to CPD, TKD is more friendly for the training process due to its capability of capturing complex data patterns using multi-dimensional factors~\cite{tucker-cp-comp}.
Also, TKD involves deterministic tensor dimensions directly controlled by the choice of a parameter, pruned rank.
Therefore, we focus on TKD as our methodology and discuss technical details of TKD next.
%
%
%
%
%

%

\betterparagraph{TKD Formulation}
TKD decomposes a tensor into a smaller core tensor and a set of matrices equal to the order of the tensor, as illustrated in~\autoref{fig:Tucker_Decomposition}.
For a \(3^{rd}\) order tensor \(T \in \mathbb{R}^{n_1 \times n_2 \times n_3}\), Tucker Decomposition can be summarized as the following:
\[T \approx \Gamma\times_1U^{1}\times_2U^{2}\times_3U^{3} = K \]
where \(\Gamma \in \mathbb{R}^{r_1 \times r_2 \times r_3}\) and $U^i \in \mathbb{R}^{r_i \times n_i}$ for $i=1,2,3$.
%

In the formula, \(r_1,~r_2,~r_3\) represent the decomposition ranks of a 3D tensor \(T\) (i.e., the rank after decomposition and pruning), and the tensor K represent the approximation of the 3D tensor $T$ obtained by the decomposition.
%
%
%

\betterparagraph{$i$-mode Product in TKD}
The \(i\)-mode product refers to the multiplication of core tensor with the i-th factor matrix, which is performed in the process of reconstructing the original tensor dimension.
The notation for $i$-mode product is $\times_{i}$.
Using this notation, the core tensor \(\Gamma\) and the factor matrices \(U^{1,~2,~3}\) for the $3^{rd}$ order tensor $T$ (i.e.,  $i = 1,~2,~3$) are defined as follows: 

$$(\Gamma \times_1 U^1)(n_1,r_2,r_3) = \sum_{i_1=1}^{r_1} \Gamma(i_1,r_2,r_3)U^1(i_1,n_1)$$
$$(\Gamma \times_2 U^2)(r_1,n_2,r_3) = \sum_{i_2=1}^{r_2} \Gamma(r_1,i_2,r_3)U^2(i_2,n_2)$$
$$(\Gamma \times_3 U^3)(r_1,r_2,n_3) = \sum_{i_3=1}^{r_3} \Gamma(r_1,r_2,i_3)U^3(i_3,n_3)$$

\betterparagraph{TKD Objective Function}
Since tensor decomposition approximates the original tensor, the error between the original tensor \(T\) and the reconstructed tensor \(K\) depends on the decomposition rank \(r_1,~r_2,~r_3\).
For a given set of decomposition ranks, the relative error between the original and the reconstructed tensors satisfies
$$\norm{T - (\Gamma\times_1U^{1}\times_2U^{2}\times_3U^{3})} \leq \epsilon \norm{T}$$
where $\norm{T}$ is the \textit{norm} of $T$.
The goal of Tucker-decomposition is to minimize $\epsilon$, and it can be formulated as
$$ \argmin_{\Gamma,U^1,U^2,U^3} \norm{T - (\Gamma\times_1U^{1}\times_2U^{2}\times_3U^{3})} $$
\input{algorithms/tucker_algorithm}

\revision{
\betterparagraph{TKD Algorithm} \autoref{alg:tucker_algo} describes Higher-Order Orthogonal Iteration (HOOI), an iterative algorithm computing a core tensor and factor matrices of a tensor$T$~\cite{de2000multilinear}.
The algorithm keeps performing singular value decomposition (SVD) for each factor matrix until the convergence criteria are met.
%
%
%
Generally, a higher decomposition rank results in a better approximation.
While the lower bound of \(r_1,~r_2,~r_3\) is \(1\), the upper bound is usually taken as \(r_i = n_i,~i = 1,2,3\) for a good approximation. In our experiments, we prune the decomposition rank \(r_1 = r_2 = r_3 \in [1, \min(n_1,n_2,n_3)]\) and use L2-norm for the algorithm. 
}

\betterparagraph{Model Compression using TKD}
To effectively compress the model, the ranks after low-rank pruning, and pruned rank (PR), need to be carefully selected. 
We formulate the maximum PR that makes the resulting model smaller than the original model.
First, we compute the number of parameters before and after the decomposition and the compression ratio using them.
$$\# ~parameters~before~decomposition = H \times W$$
$$\#~parameters~after~decomposition = H \times PR + PR \times PR + PR \times W$$
$$Compression~Ratio = \frac{H \times W}{H \times PR + PR \times PR + PR \times W}$$
The formula indicates that the smaller the PR is (preferably PR $\ll$ min(H, W)), the higher the compression ratio is.
%

To achieve the compression effect, the compression ratio needs to be larger than 1, which can be represented as follows:
$$Compression~Ratio > 1$$
$$\frac{H \times W}{H \times PR + PR \times PR + PR \times W} > 1$$
Rearranging terms in the above inequality, we obtain the following: $$PR^2 + (H + W) \times PR - H \times W < 0$$
Solving the above quadratic inequality for PR, the results indicate the upper-bound of the pruned rank that can compress the model, which is presented as follows:
$$PR < (\frac{\sqrt{(H+W)^2 + 4\times H\times W} - (H + W)}{2})$$
Note that the inequality above indicates the upper-bound value of PR for compression.
As discussed, lower PR leads to a better compression ratio.
Therefore, minimizing PR while maintaining the model performance is one of the key optimizations needed for TKD.

The compression with TKD with a proper PR can also enhance the OI.
For example, after decomposing entire weights in five layers of Llama2-7B with the PR of 1 (15\% parameter reduction in ~\autoref{tab:layer_decomposed}).
In that case, assuming the base processing style (reconstructing the original weight tensor and computing the linear layer), OI changes from 63 to 74.26, a 17.8
The increased OI is based on the changes on both numerator (number of operations) and denominator (memory traffic) in~\autoref{eq:op_intensity}: 0.12\% increment in the number of operations for reconstructing weight tensors with 15\% model size reduction: (100 + 0.12) / (100 - 15) = 1.178.
On the machine with 4 $\times$ NVIDIA A100 80GB PCIe GPUs used for the characterization and case studies (~\autoref{subsec:space_characterization}), the enhanced OI in combination of our optimizations lead to 8.5\% latency reduction, on average across all benchmarks.


%

%
%
%
%


%
%
%

\subsection{Problem Statement}
\label{subsec:problem_statement}

%
%
%

As discussed in~\autoref{subsec:challenges_for_llms}, LMs are memory-bound, which motivates us to explore optimization techniques that enhance the operational intensity by reducing memory bandwidth requirements.
TKD with rank-pruning is a promising approach for that as discussed in~\autoref{subsec:bg_tucker_decomposition}, but the approach involves a major trade-off between the model performance (e.g., accuracy) and computational performance (e.g., latency) due to the nature of rank pruning, which is not well-explored yet.

Therefore, to shed light on the trade-off of TKD, we aim to (1) \textit{profile the accuracy-computational performance trade-off space of TKD on recent LMs and report the insights from the results}.
Also, to enable systematic trade-off space exploration, we aim to (2) \textit{formulate the TKD design choices and space}.
Finally, we aim to (3) \textit{identify effective design choices of TKD}, which can be useful insights for effective TKD trade-off space exploration.

%% file: algorithms/tucker_algorithm.tex
\RestyleAlgo{ruled}
\SetKwComment{Comment}{/* }{ */}
\SetKwInOut{KwIn}{Input}
\SetKwInOut{KwOut}{Output}

\begin{algorithm}[hbt!]
\small
\caption{\revision{Tucker Decomposition via Higher-Order Orthogonal Iteration (HOOI)}}
\label{alg:tucker_algo}
\KwIn{input tensor $T$, decomposition rank ($r_1,~r_2,~r_3$), tolerance $\tau$, max iterations $iter_{max}$}
\KwOut{core tensor $\Gamma$, factor matrices $U^1,~U^2,~U^3$}

Initialize $U^2,~U^3$ with orthonormal columns

num\_iters = 0

norm\_error = $\infty$

\While{num\_iters < $iter_{max}$ and norm\_error > $\tau$} 
{
    $P = T \times_2 (U^2)^T \times_3 (U^3)^T$
    
    $U^1 = SVD(r_1, P_{(1)})$
    
    $Q = T \times_1 (U^1)^T \times_3 (U^3)^T$
    
    $U^2 = SVD(r_2, Q_{(2)})$
    
    $R = T \times_1 (U^1)^T \times_2 (U^2)^T$
    
    $U^3 = SVD(r_3, R_{(3)})$
    
    num\_iters += 1
    
    norm\_error = $\norm{ T - (\Gamma \times_1 U^{1} \times_2 U^{2} \times_3 U^{3})}$
}

$\Gamma = T \times_1 (U^1)^T \times_2 (U^2)^T \times_3 (U^3)^T$

Return $\Gamma,~U^1,~U^2,~U^3$

\Comment{A = SVD(k, B) refers to the $k^{th}$-order truncated SVD of $B$ and set $A = [a_1,a_2,...,a_k]$, where $a_1, a_2,...,a_k$ are the $k$ largest left singular vectors of $B$}
\end{algorithm}

%% file: sections/03_Methodology.tex
\section{Decomposition Design Space Formalization and Characterization}
\label{sec:methodology}

\insertFigure{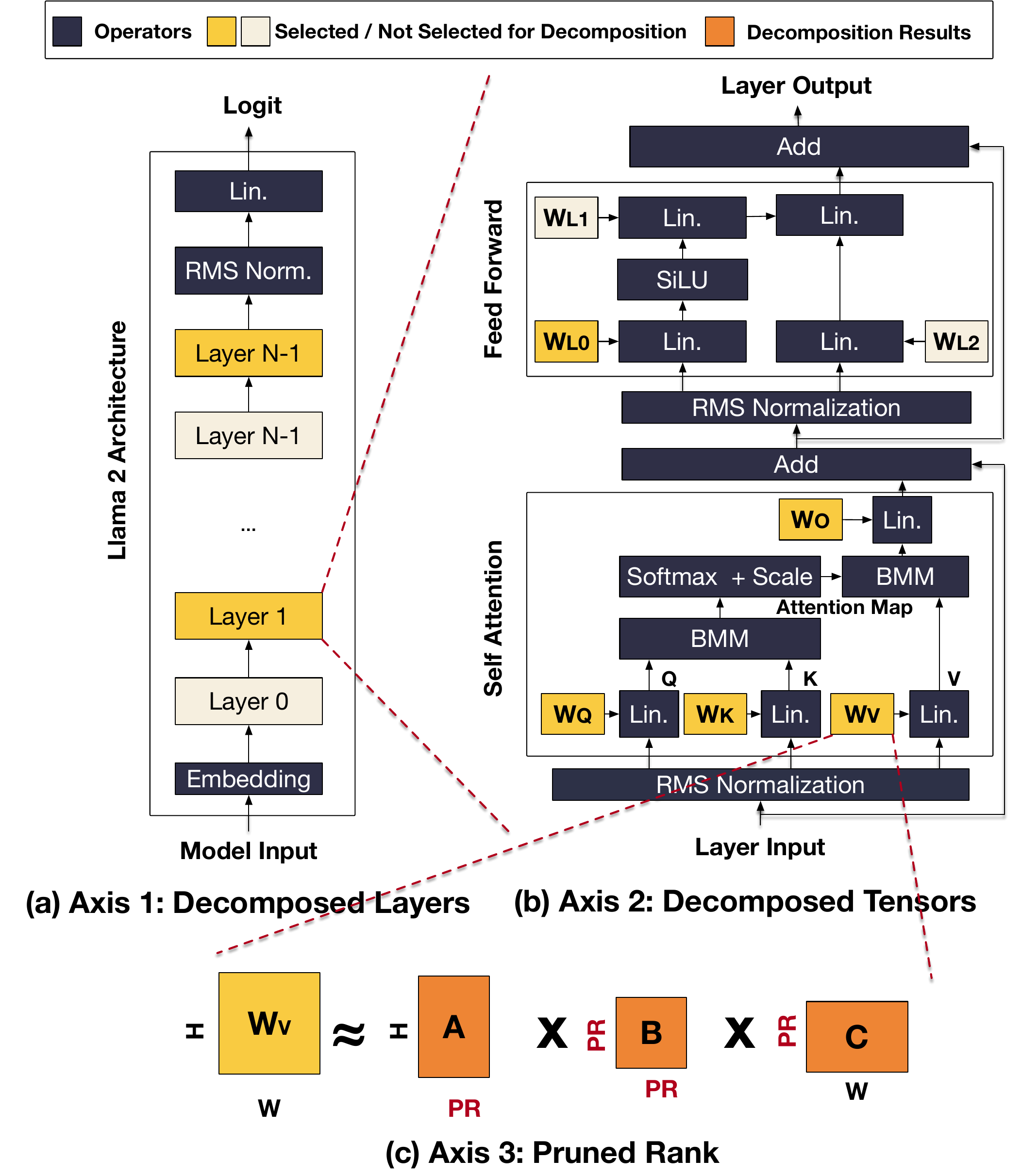}{An illustration of three axes of the decomposition configurations discussed in~\autoref{def:decomp_config}. (a): Choice of the layers to decompose, (b): Choice of tensors within each layer to decompose, (c): The choice of pruned rank (PR) to be used for each decomposed tensor. "Lin." refers to the linear layer. }

We apply TKD to the weights of pre-trained LMs and decompose weights for linear layers into three smaller matrix multiplications.
We reduce the model size by choosing a pruned rank significantly smaller than the dimension of the weight tensor (e.g., the pruned rank of 1 out of 4096 in Llama2-7B~\cite{touvron2023llama}).
We also choose to gradually compute the decomposed linear layer without reconstructing the weight tensor to minimize the number of FLOPs for the same chained matrix multiplication originating from the decomposition\footnote{Note that matrix multiplication is associative, and FLOPs for chained matrix multiplication varies depending on the compute order with associativity}, considering the dimensionality we find in LLMs such as Llama-2-7B and our choice of small PR.

%
Although Tucker decomposition is beneficial for the latency and model size reduction, we need to consider the impact on the model's accuracy when applying Tucker decomposition. 
That is, the model accuracy degradation needs to be confined within an acceptable range while maximizing latency and energy benefits.
To summarize the optimization goal, we formalize the goal assuming equal importance on latency and energy (i.e., targeting energy-delay-product as the objective metric) as follows:

\begin{definition}
\textsc{Design Goal of low-rank decomposition} \\\
Given an accuracy threshold $\tau$, find a decomposition configuration $\gamma$ such that 
$$\argmin_{\gamma: max(Accuracy_{Original} - Accuracy(\gamma),0) < \tau } Latency(\gamma) \times Energy(\gamma)$$  
\label{def:design_goal}
\end{definition}

In~\autoref{def:design_goal}, $\tau$ refers to an accuracy drop tolerance (e.g., 4\%).
The optimized variable, $\gamma$, refers to how we decompose a language model, which consists of three choices illustrated in~\autoref{fig:DesignSpace}.
(We discuss the formal definition of $\gamma$ later in~\autoref{def:decomp_config}.)
$\gamma$ enables us to describe any partial decomposition configuration of a model, which helps to navigate the accuracy-efficiency trade-off.
\revision{Note that the above formulation does not reject a decomposition configuration that may enhance the resulting accuracy after decomposition.}

However, because recent language models are complex and large, the number of possible ways of decomposition (i.e., decomposition design space) is massive (e.g., $\mathbf{O(2^{39})}$ ways for the smallest Llama 2 variant, Llama2 - 7B).
Therefore, we first formalize the decomposition design space and characterize it to identify effective and ineffective decomposition configurations.
Based on the characterization results, we prune the ineffective decomposition configurations and reduce the decomposition design space to a tractable size (e.g., for an LLM, Llama2-7B, the design space is reduced from O($2^{39}$) to O($2^{5}$).

\subsection{Decomposition Space}
\label{subsec:decomp_space_formulation}

Because of the uniformity of the building blocks (i.e., the same block is repeated) in language models, as illustrated in~\autoref{fig:DesignSpace} (a), we target homogeneous decomposition schemes for each layer.
That is, we prune the same number of ranks and select the same set of weight tensors to be decomposed within each layer.
Combined with those two choices (number of ranks after pruning and tensors to be decomposed), describing the layers to be decomposed provides a complete description of one decomposition configuration.
We formalize the decomposition configuration and design space next to provide a precise definition of them.

\begin{definition}
\textsc{Decomposed Layers and Tensors} \\\
For a given model $m$, which has $N_{Layers}(m)$ layers and  $N_{Tensors}(m)$ of decomposable weight tensors in each layer, the choices of decomposed layers $Decomp_{Layers}(m)$) and tensors ($Decomp_{Tensors}(m)$) are defined as follows:
$$Decomp_{Layers}(m) = \{DL_{0}, DL_{1}, ... DL_{L}\} $$
$$Decomp_{Tensors}(m) = \{DT_{0}, DT_{1}, ... DT_{K}\} $$
where 
$$ (L,K \in \mathbb{Z}) \land (0 \leq L < N_{Layers}(m)) \land (0 \leq K < N_{Tensors}(m))
\square $$
\label{def:base_definition}
\end{definition}

In~\autoref{def:base_definition}, we describe the choice of decomposed layers and tensors as sets of corresponding layers and tensor IDs represented in integers.
If the K and L are set to zeros, corresponding $Decomp_{Layers}$ and $Decomp_{Tensors}$ become empty sets, which expresses the original model without any decomposition.

\begin{definition}
\textsc{Pruned ranks} \\\
For a given model $m$, the rank after pruning (or pruned ranks), $PR(m)$ is defined as follows:
\begin{equation*}
\small
    \begin{aligned}
        PR(m) = \{(l,k,p) ~|~ (l,k,p \in \mathbb{Z}) \land (0 \leq k < N_{Tensors}(m))\\
        \land (0 \leq l < N_{Layers}(m)) \land (0 < p \leq rank(l,k)) \}
    \end{aligned}
\end{equation*}
where $rank(l,k)$ refers to the rank of a weight tensor $k$ in layer $l$. $\square$
\label{def:pruned_ranks}
\vspace{5mm}
\end{definition}

The formulation in~\autoref{def:pruned_ranks} indicates that the pruned rank cannot exceed the original rank.
Also, ~\autoref{def:pruned_ranks} allows us to describe the decomposition without rank pruning by setting the pruned rank ($p$) the same as the original rank.
Using ~\autoref{def:base_definition} and ~\autoref{def:pruned_ranks}, we define a complete low-rank decomposition configuration as follows. 

\begin{definition}
\textsc{Low-rank Decomposition Configuration ($\gamma$)} \\\
For a given model $m$, which has $N_{Layers}(m)$ layers, $N_{Tensors}(m)$ of decomposable weight tensors in each layer, and $Dim(m,ID_{Layer}, ID_{Tensor})$ dimensions, a decomposition configuration for model $m$ ($\gamma(m)$) is defined as follows:
{\small
$$ \gamma(m) = (PR(m), Decomp_{Layers}(m), Decomp_{Tensors}(m)) \square$$
}
\label{def:decomp_config}
\end{definition}

In~\autoref{def:decomp_config}, we define the decomposition configuration as a tuple of $PR(m)$, $Decomp_{Layers}$, and $Decomp_{Tensors}$, which are defined in~\autoref{def:pruned_ranks} and~\autoref{def:base_definition}.
The tuple captures the three major decomposition axes illustrated in~\autoref{fig:DesignSpace}.
Before we define the low-rank decomposition design space, we first define the validity of a decomposition configuration.

\begin{proposition}
\textsc{Validity of a Decomposition Configuration ($Val(\gamma)$)} \\\
For a given model $m$, which has $N_{Layers}(m)$ layers, $N_{Tensors}(m)$ of decomposable weight tensors in each layer, and $Dim_{Min}(m)$ to be the smallest weight matrix dimension (i.e., number of columns), a decomposition configuration for model $m$ \\
($\gamma(m) = (PR(m), Decomp_{Layers}(m), Decomp_{Tensors}(m))$), $\gamma(m)$ is valid if the following conditions are met:
\begin{equation*}
\small
\begin{aligned}
\forall (l,k,p) \in PR(m), l \in Decomp_{Layers}(m) \land k \in Decomp_{Tensors}(m) \\
\land |PR(m)| = (|Decomp_{Layers}| - 1) \times (|Decomp_{Tensors}| - 1) + 1 \square
\end{aligned}
\end{equation*}
\label{prop:config_validity}
\end{proposition}

Because the individual validity of $PR$, $Decomp_{Layers}$, and $Decomp_{Tensors}$ are checked in their definitions in~\autoref{def:pruned_ranks} and~\autoref{def:base_definition}, we need to check the validity as their combination.
Note that $Decomp_{Layers}$ and $Decomp_{Tensors}$ are independent, based on their definition in~\autoref{def:base_definition} (i.e., selection of the decomposed layers and tensors within each decomposed layer is independent).
However, the definition of the pruned ranks,~\autoref{def:pruned_ranks}, contains the layer and tensor IDs.
Therefore, we need to ensure the pruned ranks cover all the decomposed (layer, tensor) combinations, and ~\autoref{prop:config_validity} states that condition.

Using the definitions and proposition, we can define the decomposition design space as follows.

\begin{definition}
\textsc{Low-rank Decomposition Design Space ($S_{LR}$)} \\\
For a given model $m$, the decomposition design space($S_{LR}(m)$) is defined as follows:
{\small
$$S_{LR}(m) = \{\gamma_{i} ~|~ Val(\gamma_{i}) \land i \in \mathbb{Z} \land i \geq 0 \} \square$$
}
\label{def:design_space}
\end{definition}

\autoref{def:design_space} states that the decomposition design space is a set of all valid decomposition configurations.
Using the observation, we analyze the scale of the design space using the Big-O notation as follows:

\begin{theorem}
\textsc{Decomposition Design Space Size ($|S_{LR}|$)} \\\
For a given model $m$ and its decomposition design space($S_{LR}(m)$), 
{\small 
$$|S_{LR}(m)| = (2^{N_{Tensors}(m)} - 1) \times (2^{N_{Layers}(m)} -1) \times rank(l,k) + 1$$
}
\begin{proof}
The size of the decomposition design space $|S_{LR}(m)|$ is the number of elements within $S_{LR}(m)$. Because $S_{LR}(m)$ is a set of valid decomposition configurations $\gamma$, we count all the possible $\gamma = (PR, Decomp_{Layers}, Decomp_{Tensors})$. \\
\textbf{(1) The number of possible choices for pruned ranks ($|PR|$)} \\ 
Based on~\autoref{def:pruned_ranks}, 
$ \forall (l,k,p) \in PR(m), p \leq rank(l,k)$. That is, the number of available choices for $p$ is $rank(l,k)$. \\
\textbf{(2) The number of possible choices for decomposed layers} \\
We can select to decompose 0 to $N_{Layers}(m)$ layers. However, in addition to the total number of decomposed layers, we need to specify which layers are decomposed (e.g., decomposing two layers, we can select layers 0 and 1, 1 and 2, 0 and 3, etc.), which can be represented using combinations. Therefore, the number of possible choices is as follows:
$$ \Sigma_{l=0}^{N_{Layers}(m)} {N_{Layers}(m) \choose l}$$
\textbf{(3) The number of possible choices for decomposed tensors} \\
Following the same method as (2), the number of possible choices is represented as follows:
$$ \Sigma_{k=0}^{N_{Tensors}(m)} {N_{Tensors}(m) \choose k}$$\\
\\
\textbf{(4) The number of valid combinations of decomposed layers and tensors} \\
The number of all possible combinations of decomposed layers ($N_{AllComb}$) and tensors can be counted as $(2) \times (3)$:
{\small
$$N_{AllComb} = \Sigma_{l=0}^{N_{Layers}(m)} {N_{Layers}(m) \choose l} \Sigma_{k=0}^{N_{Tensors}(m)} {N_{Tensors}(m) \choose k}$$
}
However, the above equation counts the cases where the model is not decomposed ($l=0$ or $k=0$) multiple times, while they should only be counted once. Therefore, all the valid combinations of decomposed layers and tensors ($N_{ValComb}$) are counted as:
{\small
\begin{align*}
N_{ValComb} = \Sigma_{l=1}^{N_{Layers}(m)} {N_{Layers}(m) \choose l} \Sigma_{k=1}^{N_{Tensors}(m)} {N_{Tensors}(m) \choose k} + 1
\end{align*}
}
Because we count all the possible and valid $\gamma = (PR, Decomp_{Layers}, Decomp_{Tensors})$ (i.e., all the possible combinations of (1) and (4) ), 
{\small
\begin{align*}
|S_{LR}&(m)| = (4) \times (1) \\
            &= \left(\Sigma_{l=1}^{N_{Layers}(m)} {N_{Layers}(m) \choose l} \Sigma_{k=1}^{N_{Tensors}(m)} {N_{Tensors}(m) \choose k} + 1\right) \\ 
            & \times rank(l,k) \\ 
            &= (2^{N_{Layers}(m)} - 1) \times (2^{N_{Tensors}(m)} - 1) \times rank(l,k) + 1
\end{align*}
}
where $rank(l,k)$ is the target pruned rank for a uniform decomposition of all tensors.\\
\end{proof}
\label{thrm:design_space_size}
\end{theorem}

\input{tables/tab_design_space}

Based on~\autoref{thrm:design_space_size}, we can estimate the size of the decomposition space as O($2^{N_{Layers}(m) + N_{Tensors}(m)}$), even if we apply uniform decomposition as assumed in this subsection.
Using this notation, we can analyze the design space of recent language models in~\autoref{tab:design_space_size}.
With the data, we observe the decomposition design space size for large language models such as Llama2~\cite{touvron2023llama} is intractably huge.
Therefore, we characterize design space to identify ineffective decomposition configurations and prune the space.

\subsection{Decomposition Space Characterization}
\label{subsec:space_characterization}

To explore decomposition design space pruning opportunities, we investigate the impact of each decomposition axis: (1) pruned rank, (2) tensors, and (3) layers.

\betterparagraph{Characterization Methodology}
We use LLama2-7B~\cite{touvron2023llama} for the characterization studies.
We run six benchmarks listed in~\autoref{tab:benchmarks} and track how accuracy changes when we individually change each axis of the design space individually.
We run the workloads on a system with AMD EPYC 7763 processor, 1TB of main memory, and four NVIDIA A100 GPUs with 80GB HBM2e memory for each.
To measure the latency of the multi-GPU system, we measure GPU runtime utilizing \textit{torch.cuda.event} APIs.
We also utilize NVIDIA's System Management Interface (\textit{nvidia-smi}) to measure power consumption and memory usage.
We calculate the area under the power-time graph using \textit{nvidia-smi}-reported average power information to estimate the GPU energy consumption.

\insertWideFigure{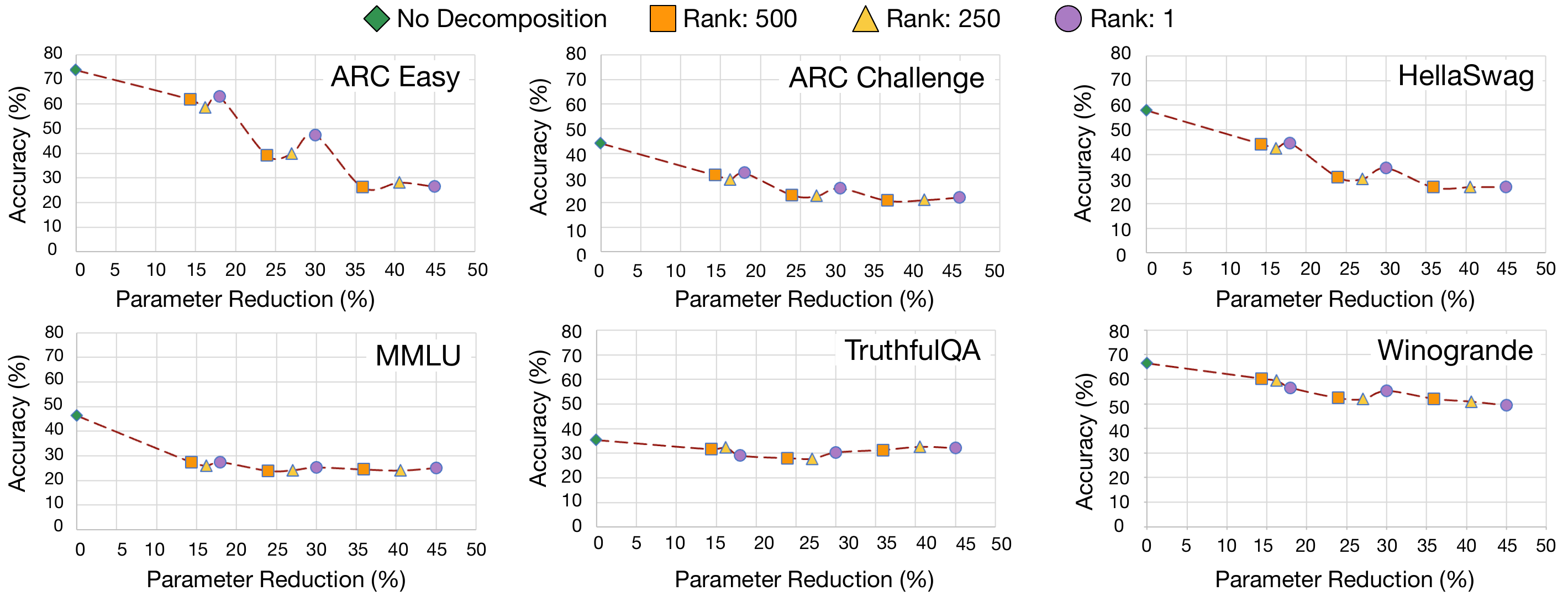}{Impact of Rank on Accuracy. We prune ranks from the original (4096) to 500, 250, and 1. By pruned rank (PR), we refer to the remaining rank after rank pruning. The accuracy with no decomposition is based on the reported accuracy in the original Llama2 publication~\cite{touvron2023llama}.}


\betterparagraph{Choice of Pruned Rank}
We decompose all the tensors illustrated in~\autoref{fig:DesignSpace}(b) in each layer, varying the pruned ranks to observe the impact of pruned rank choices on accuracy.
We select different combinations of layers as shown in~\autoref{tab:layer_decomposed} to observe the trend on different parameter reduction rates.
We choose three values of pruned ranks for our characterization: 1, 250, and 500.
While the original ranks of BERT and Llama-2-7B are 768 and 4096, respectively, we do not increase the pruned rank above 500 to ensure a meaningful model size reduction rate.

\subsection{Observation from the Characterization}
\label{subsec:observation}
In~\autoref{fig:Rank_Sensitivity_Graph}, we observe that the pruned rank has minimal effect on the accuracy score compared.
Across all six benchmarks, we notice consistent patterns, indicating an average accuracy fluctuation of 1.5\%.
Based on this observation, we select rank-1 decomposition for the main case study because it provides the highest model size reduction.

\subsubsection{The impact of Decomposed Tensor Choices}
\label{subsubsec:tensor_impact}

\insertFigure{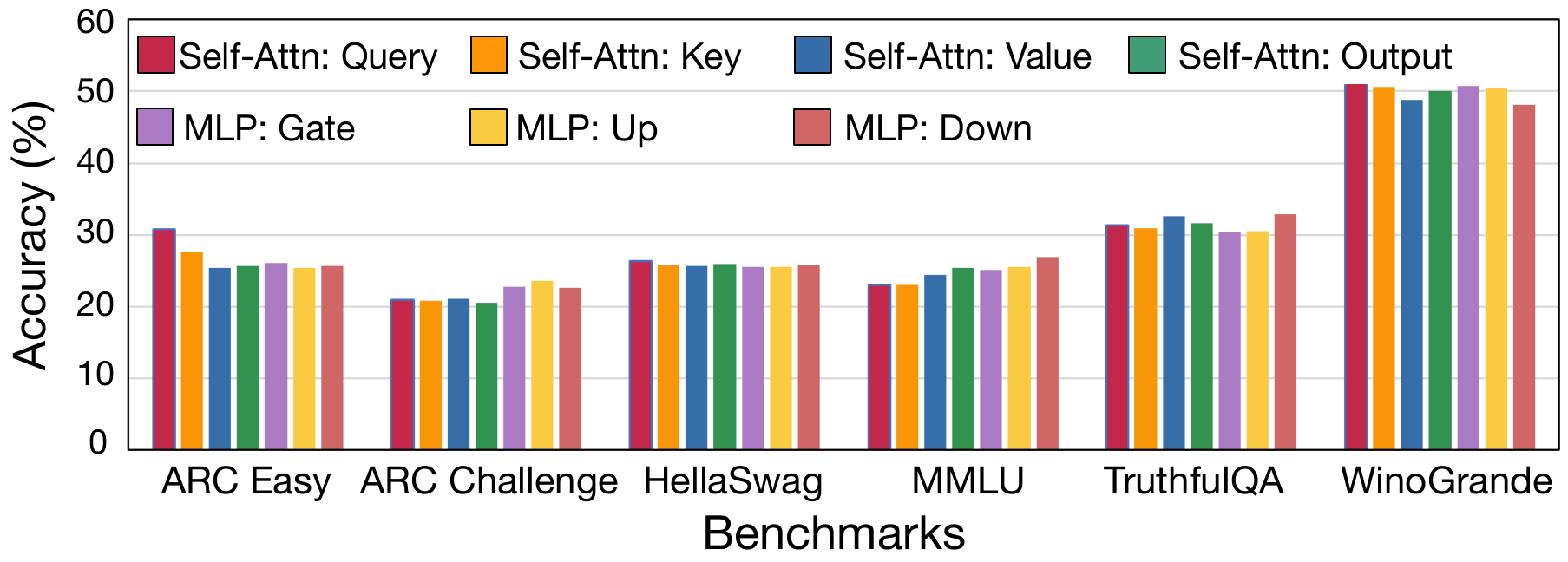}{The impact of decomposed tensor choices on the accuracy on Llama2-7B. We decompose one tensor in all the decoder layers.}
\insertFigure{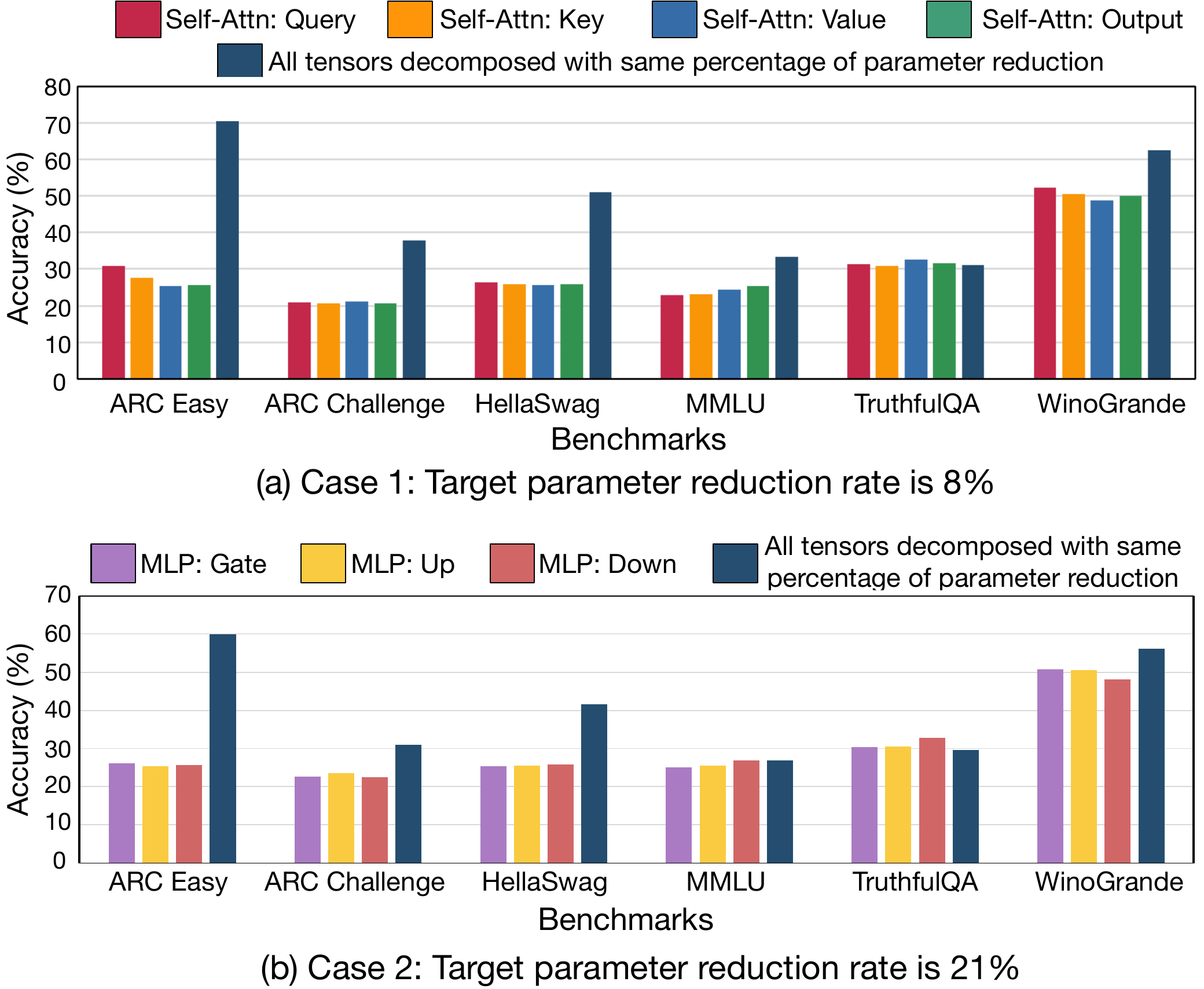}{To understand the impact of decomposed tensor choices, we compare two different ways for achieving similar parameter reduction rates: (1) decompose a specific tensor in many layers (2) decompose all tensors and select less number of layers to be decomposed. The right-most black bar corresponds to (2), and all the other bars correspond to (1). We compare those two approaches using two different parameter reduction targets: (a) 8\% and (b) 21\%.}
As shown in~\autoref{fig:ModelArchitecture}, each BERT layer contains six weight tensors: Query ($W_Q$), key ($W_K$), and value ($W_V$) projection, self-attention output projection ($W_{SO}$), intermediate fully-connected layer ($W_{Int}$), and output fully connected layer ($W_O$). 
Each Llama-2-7B layer contains seven weight tensors: Query ($W_Q$), key ($W_K$), value ($W_V$) projection, self-attention output projection ($W_{SO}$), Multi-Layer Perceptron (MLP) gate projection ($W_G$), MLP up projection ($W_U$) and MLP down projection ($W_D$).
We first decompose each tensor separately to analyze their individual impact on the overall model accuracy.
We also decompose multiple tensors in various combinations (i.e., decomposed tensor "group") to analyze the interplay of decomposed tensors.
We present the results in figures \ref{fig:Tensor_Sensitivity} and \ref{fig:Tesor_Sensitivity_grouped}, and observe the following:

\betterparagraph{Observation 1. All the tensors in self-attention and MLP modules are equally sensitive to decomposition when compared within each group of decomposed tensors, but individual sensitivity varies across different combinations of decomposed tensors.}
We decompose individual tensor in Llama-2-7B layers using the pruned rank of 1 in one or all the decoder layers and present the accuracy results in~\autoref{fig:Tensor_Sensitivity}.
From the results, we observe that the choice of the tensor overall does not result in significant differences in the resulting accuracy, which indicates that the accuracy is not sensitive to the choice of tensors.
%
%

\betterparagraph{Observation 2. When targeting the same model size reduction rate, the accuracy is more sensitive to the number of layers containing decomposed tensors (i.e., \# of decomposition target layers) than the number of decomposed tensors in decomposed layers} 
%
Targeting the same (or similar) model size reduction rate, we have multiple potential decomposition configurations, which adjust the number of decomposed layers and tensors.
For example, one decomposition configuration that decomposes one tensor in 30 decoder layers may achieve the same model size reduction as another configuration that decomposes three tensors in 10 layers to achieve .
To guide the right choice of decomposition configurations, we characterize the accuracy under various configurations with the same target parameter reduction rate and present the results in~\autoref{fig:Tesor_Sensitivity_grouped}. 

When we decompose the Query tensor in all the 32 decoder layers of Llama-2-7B, we achieve a parameter reduction of \(8\%\). 
However, that approach led to more than 50\%p\footnote{We use "percentage point" (\%p) to avoid confusion. "percentage point" refers to the absolute difference in two percentage numbers. For example, 47\% is 5\%p higher than 42\% and (0.47/0.42-1)= 11.2\% higher than 42\%.} of accuracy loss while it only provided \(8\%\) parameter reduction.
In contrast, if we decompose all the tensors and reduce the number of decomposed layers to keep the overall parameter reduction ratio identical, we observe a much smaller accuracy drop (3\%p).
%
%
We observe similar trends across those two cases and for BERT as well.
The results indicate that we should decompose more tensors rather than decomposing more layers with the same parameter reduction rate.

\subsubsection{The Impact of Decomposed Layer Choices}
\label{subsubsec:layer_impact}

\insertFigure{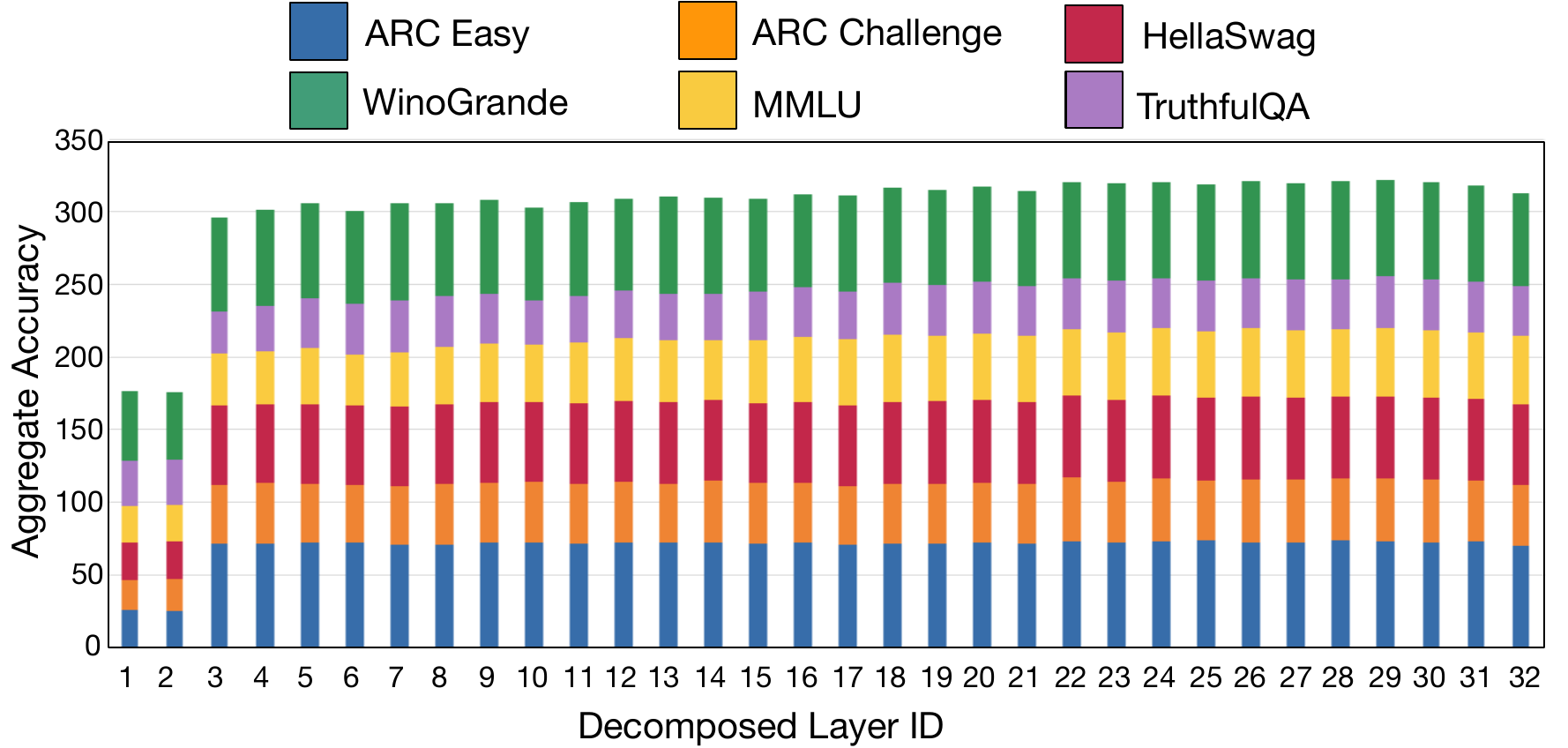}{The aggregate accuracy across six benchmarks when we decompose different layers of Llama2-7B. We select one layer to be decomposed and plot the correlation between the location of the selected layer and accuracy.}

\insertFigure{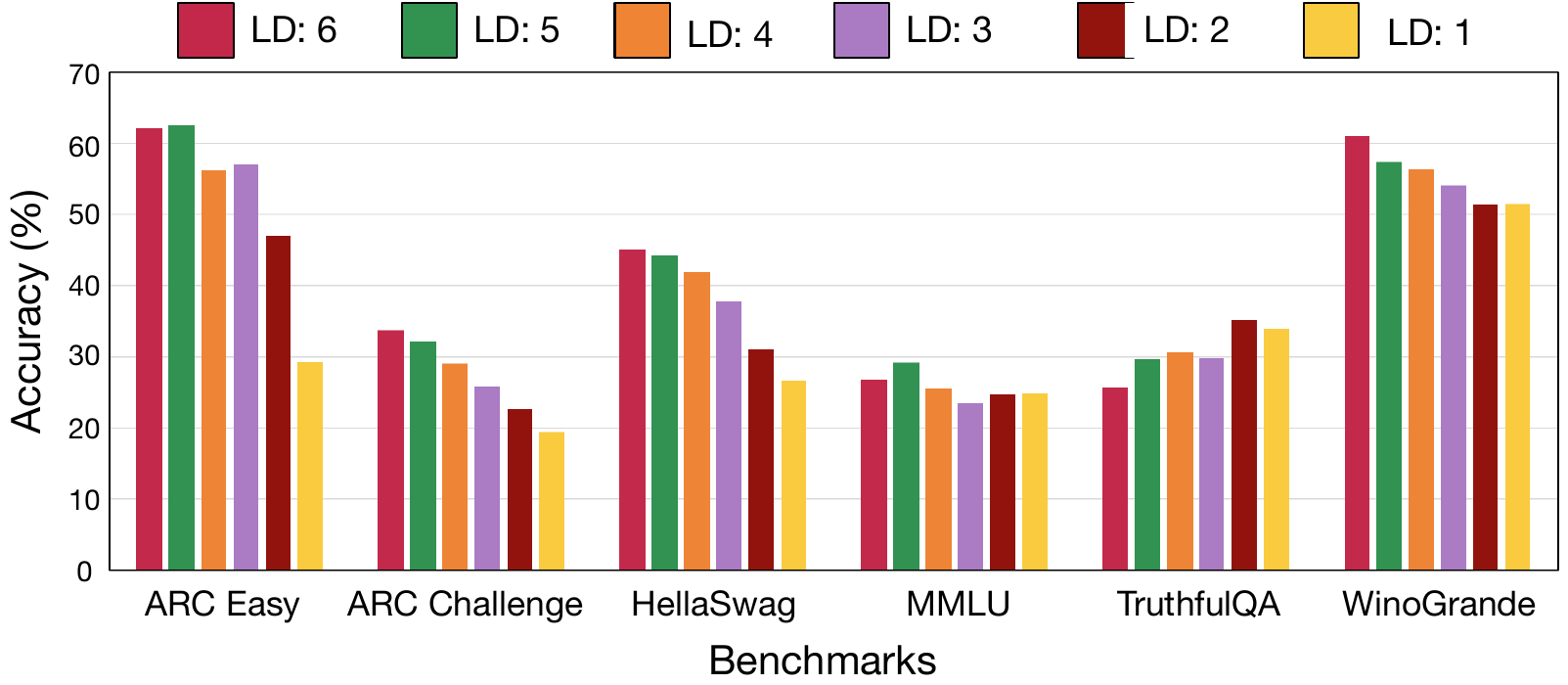}{The accuracy results on Llama-2-7B when applying different distances between decomposed layers. We decompose five layers starting from layer 3 with uniform distance between decomposed layers. (LD: Layer Distance)}

We decompose each encoder layer in BERT and decoder layer in Llama-2-7B, respectively, and characterize the resulting accuracy.
We show the Llama-2-7B results in~\autoref{fig:Layer_Sensitivity} with aggregate accuracy across all benchmark to show the overall impact and break-down into each benchmark.

\betterparagraph{Observation 3. Decomposing early layers significantly reduces the accuracy}
From the results, we observe that decomposing any of the first two layers results in significant accuracy degradation, reducing 43\% of accuracy compared to the model without decomposition, on average.
Decomposing one of the last three layers also reduce the accuracy, but the degree is minor, 3\%, on average.
%

\betterparagraph{Observation 4. Decomposing close layers significantly degrades the accuracy}
We extend our characterization to the multiple decomposed layer cases where we select five decomposed layers with uniform distance starting from layer 3.
For example, if the distance is 5, layers 3, 8, 13, 18, and 23 are decomposed.
%
%
We present the characterized accuracy in~\autoref{fig:Layer_Distance_Sensitivity}.
From the results, we first observe that decomposing close layers significant reduces the accuracy for all benchmarks except TruthfulQA.
For example, the accuracy of Llama-2-7B running ARC Easy reduces from 62\% to 29\% when changing the distance between decomposed layers decreases from 6 to 1 (i.e., decomposed layers are adjacent).
Only TruthfulQA results in the opposite trend, and this can be understood from the difference in the task; as listed in~\autoref{tab:benchmarks}, only TruthfulQA checks the truthfulness of the answer (focusing on the reliability of LMs) while other benchmarks mainly focus on the reasoning within given contexts.
This indicates that the desired distance between decomposed layers depends on the downstream tasks.
%
%
%

\subsection{Insights from the Characterization}
\label{subsec:characterization_insights}

Based on the analysis presented in this section, we can draw the following conclusions:
\begin{itemize}
{\item Rank-1 decomposition seems to produce similar results to higher-rank decomposition. Therefore, it is better to use rank-1 decomposition as it achieves higher parameter reduction with the same loss of accuracy as other higher-rank decompositions.}
{\item Decomposing all the tensors within fewer encoder/decoder layers or self-attention/MLP modules is better than decomposing the same number of tensors (but not all) across more layers or modules. 
}
{\item It is best to avoid decomposing early and late layers because the accuracy is highly sensitive on the decomposition of such layers.}
{\item For most cases, decomposing close layers significantly reduces the accuracy. However, the trend can depend on the task of the target benchmark.
}
\end{itemize}
%

%


%% file: tables/tab_design_space.tex
\begin{table}[]
\caption{A summary of the model parameters and corresponding design space size of recent language models. NDT refers to the number of decomposable tensors in each layer.}
\label{tab:design_space_size}
\centering
\begin{tabular}{|l|r|r|r|}
\cline{1-4}
\multicolumn{1}{|c|}{\textbf{Model}} 
& \multicolumn{1}{c|}{\textbf{\# Layers}} 
& \textbf{NDT}
& \textbf{\begin{tabular}[c]{@{}l@{}}Decomposition \\ Design Space\end{tabular}} \\
\hline
Bert-Base
& 12
& 6
& O($2^{18}$) \\
\hline
Bert-Large
& 24
& 6
& O($2^{30}$) \\
\hline
Llama 2 - 7B
& 32 
& 7                                                                    & O($2^{39}$) \\ 
\hline
Llama 2 - 70B  
& 80
& 7
& O($2^{87}$) \\ 
\hline
\end{tabular}
\end{table}

%% file: sections/04_CaseStudy.tex
\section{Case Study}
\label{sec:case_study}

We conduct an in-depth case study to characterize the trade-off space of Tucker low-rank decomposition LLMs.

\subsection{Configuration and Methodology}
\label{subsubsec:eval_settings}

\input{tables/tab_llm_benchmarks}
\input{tables/Decomposition_Configuration}
\input{tables/tab_eval_profiling_time.tex}

\betterparagraph{Model} We use Llama-2-7b-chat-hf as a representative state-of-the-art LLM.

\betterparagraph{Benchmarks}
We evaluate our low-rank decomposition method on all the standard benchmarks included in the HuggingFace Open LLM Leaderboard~\cite{huggingface2024llmleaderboard}, which includes ARC (Easy and Challenge), HellaSwag, MMLU, TruthfulQA, and WinoGrande.
These benchmarks span a variety of domains where LLMs are often utilized, which provides a good basis for testing the performance of our methodologies comprehensively.
We discuss how we utilize each benchmark for four LLM tasks as follows:

\begin{itemize}
    {\item \textbf{Commonsense Reasoning}: We report the zero-shot accuracy of ARC Easy and Challenge, HellaSwag, and WinoGrande. While the ARC and WinoGrande datasets use the question-answering format to test common sense reasoning, HellaSwag tests the sentence completion ability of a model.}
    {\item \textbf{Multitask Accuracy:} We report the 0-shot accuracy of the MMLU dataset, which tests a model's accuracy on multiple tasks like world knowledge and problem-solving.}   
    {\item \textbf{Truthfulness (safety benchmark):} We report the accuracy on the TruthfulQA benchmark, which tests if a model can generate reliable outputs that agree with factuality and common sense.}
\end{itemize}

\betterparagraph{Hardware Platform}
We use the same hardware platform (4 $\times$ NVIDIA A100 80GB PCIe) as the initial characterization study, discussed in~\autoref{subsec:space_characterization}.

\betterparagraph{Software Platform} We adopt EluetherAI's~\cite {eval-harness} \texttt{lm-evaluation-harness} framework to measure the accuracy of Llama-2-7B on the benchmarks we use.

\betterparagraph{Batch Size} 
We set the batch sizes to be the maximum size for the available GPU memory and utilize all four NVIDIA A100 GPUs in parallel.

\betterparagraph{Decomposition Configurations}
The choice of those configurations are driven by the learning from our characterization in~\autoref{subsec:characterization_insights}, which can be summarized as follows:
\textbf{(1)} apply pruned rank of 1, \textbf{(2)} decompose all tensors within selected layers for decomposition,\textbf{(3)} avoid decomposing early layers (for Llama-2-7B, we start from layer 3), and \textbf{(4)} keep the distance between decomposed layers as far as possible.
Following the methodology, we use the decomposed layers listed in~\autoref{tab:decomposition_configuration}.

\betterparagraph{Latency Measurement Methodology}
We run each benchmark for each decomposition configuration multiple times and compute the average of the latency on the machine discussed in~\autoref{subsec:space_characterization}.
We list the overall latency for each configuration (for the multiple runs) in~\autoref{tab:profiling_time}.


%

%

\insertFigure{Accuracy_Impact}{The resulting accuracy after applying low-rank Tucker decomposition in various decomposition configurations with different parameter reduction rates. }

\insertWideFigure{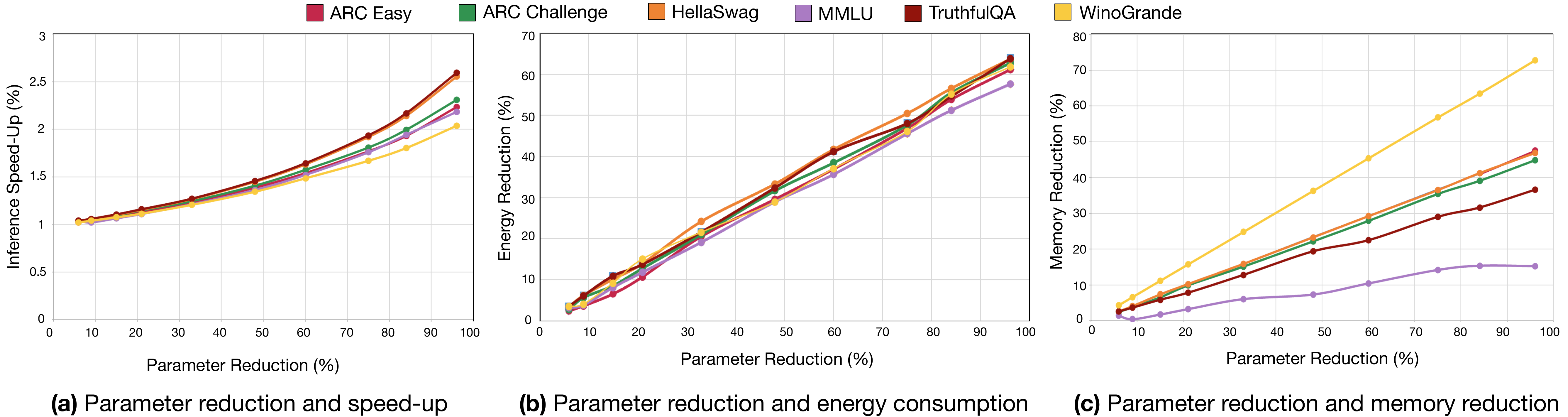}{The impact of model parameter reduction via low-rank Tucker decomposition on the speed up, energy, and memory usage.}

%
%

\subsection{Results and Discussion}
\label{subsubsec:results_and_discussion}

\betterparagraph{Accuracy} 
We plot the accuracy after low-rank Tucker decomposition in~\autoref{fig:Accuracy_Impact}, which shows the accuracy against model parameter reduction rate.
We observe that ARC Easy suffers from acute accuracy drop until the parameter reduction rate of 50\%.
In low parameter reduction (1~10\%) region, ARC Easy accuracy drops 7.4\%p for each 1\% parameter reduction. 
%
%
However, we observe that the accuracy with WinoGrande is less sensitive to the degree of decomposition.
Unlike the previous two examples, TruthfulQA presents a fluctuation in accuracy with respect to the parameter reduction, and the accuracy eventually improves to 32\% under an aggressive parameter reduction (96\%).
The results indicate the possibility that the relatively difficult (i.e., initial accuracy of the pre-trained model is relatively low; < 50\%) benchmark could be friendly to the decomposition.
%
%
%
For relatively complex benchmarks (ARC Challenge, HellaSwag, and MMLU), the accuracy degrades 4 to 11 \%p for up to a 9\% reduction in parameters.
However, the accuracy degradation becomes significant afterward, as shown in \autoref{fig:Accuracy_Impact}.
%

\betterparagraph{Inference Latency} Memory usage and traffic decrease as we reduce the model size by low-rank decomposition.
This facilitates faster inference and, consequently, less energy consumption.
In terms of the inference latency, we observe a steady reduction in inference time as model size reduces. For every \(1\%\) reduction in model size, we see an average of \(0.5\%\) reduction in inference time.

\betterparagraph{Energy}
Since LLMs are heavy workloads utilizing the GPU at 100\%, we observe that the power consumption of the GPU always stays at the maximum (300W in the case of NVIDIA A100 80GB).
Therefore, savings in inference latency result in a proportional saving in energy consumption. We observe a similar ratio as latency in energy savings where every percentage parameter reduction results in approximately \(0.5\%\) reduction in energy consumption, as shown in~\autoref{fig:Inference_Time_Reduction} (b).
Across all the experiments, the relative standard deviation of inference time and energy consumption is $2.6\%$ and $2.9\%$, respectively.

\betterparagraph{Memory Usage}
We present the memory usage reduction trend against the model parameter reduction rate in \autoref{fig:Inference_Time_Reduction} (c).
We observe a similar trend as the latency and energy: each \(1\%\) model parameter reduction leads to approximately \(0.4\%\) reduction in total memory usage of the GPU.


\subsection{Fine-tuning for accuracy improvement}
Our goal is to characterize the raw design space of low-rank decomposition on LLMs without any further fine-tuning, rather than achieving the best design point.
However, as an additional case study, we investigated the impact of fine-tuning (i.e., re-training the model with low-rank decomposed weights) for all benchmarks on selected decomposition design choices.
We follow overall the same methodology with a smaller set of configurations: the number of decomposed layers ranges from 2 to 32 (upto 33\% parameter reduction). 
The average initial accuracy on all benchmarks before decomposition is 52\%, and that after decomposition is 31\%.
Simply launching re-training of the decomposed model for two epochs, we observe that the average accuracy improves by 3.5\%p (from 31\% to 34.5\%). 
This indicates that proper fine-tuning methodology can minimize the accuracy loss of low-rank Tucker decomposition, which motivates follow-up studies.

\subsection{Insights from the Results}

Based on the results obtained in our case study, the main insights of this work are as follows:

\begin{itemize}
    {\item We show that it is possible to compress LLMs by reducing the number of parameters by up to $9\%$ with a relatively small accuracy loss of $4\%p$ to $10\%p$ depending on the benchmark, even without retraining or fine-tuning the model after decomposition.}
    {\item We also show that for every $1\%$ reduction in the model's parameters, there is a proportional decrease of $0.5\%$ in inference latency and energy consumption. The memory usage also decreases by $0.4\%$ for the same amount of parameter reduction.}
    {\item We note that a decomposed model with fewer parameters than the original model consumes considerably less energy and has lower inference latency. The inference latency and the energy consumption scale linearly with the decomposed model size, and this holds across all the benchmarks we examined.}
\end{itemize}

%% file: tables/tab_llm_benchmarks.tex
\begin{table}[]
\caption{Benchmarks used for our case studies.}
\label{tab:benchmarks}
\begin{tabular}{|l|l|r|}
\hline
\multicolumn{1}{|c|}{\textbf{Benchmark}} & \multicolumn{1}{c|}{\textbf{Task}}                                            & \multicolumn{1}{c|}{\textbf{\# of Samples}} \\ \hline 
ARC Easy           & Reasoning (Q\&A)                      & 5.2K                       \\ \hline
ARC Challenge      & Reasoning(Q\&A)              & 2.59K                      \\ \hline
HellaSwag          & Reasoning(Sentence Completion) & 10K                        \\ \hline
MMLU               & Multitask Language Understanding                         & 15.9K                      \\ \hline
TruthfulQA         & Truthfulness                                             & 1634                       \\ \hline
WinoGrande         & Reasoning (Q\&A)                   & 44K                        \\ \hline
\end{tabular}
\end{table}

%% file: tables/Decomposition_Configuration.tex
\begin{table}[]
\caption{Decomposed layer choices used in our case studies and corresponding parameter reduction rates.}
\label{tab:layer_decomposed}
\scriptsize
\centering
\begin{tabular}{|c|c|}
\hline
\label{tab:decomposition_configuration}
\textbf{\makecell{Parameter \\ Reduction (\%)}} & \textbf{Decomposed Layers}  \\ \hline
6\%                                                                    & 3, 30 \\
\hline
9\% 
& 3, 18, 32  \\
\hline
15\% 
& 3, 9, 15, 21, 27 \\
\hline
21\% 
& 5, 9, 13, 17, 21, 25, 29 \\
\hline
33\% 
& 3, 6, 9, 12, 15, 18, 21, 24, 27, 30, 32 \\
\hline
48\% 
& \begin{tabular}[c]{@{}c@{}}1, 3, 5, 7, 9, 11, 13, 15, 17, \\ 19, 21, 23, 25, 27, 29, 31\end{tabular} \\
\hline
60\% 
& \begin{tabular}[c]{@{}c@{}}2, 4, 6, 8, 10, 11, 12, 13, 14, 15, \\ 16, 17, 18, 19, 21, 23, 25, 27, 29, 31\end{tabular} \\
\hline
75\% 
& \begin{tabular}[c]{@{}c@{}}2, 4, 6, 8, 10, 11, 12, 13, 14, 15, 16,   17, 18, \\ 19, 20, 21, 22, 23, 24, 25, 26, 27, 28, 29, 30\end{tabular} \\
\hline
84\% 
& \begin{tabular}[c]{@{}c@{}}1, 3, 5, 7, 9, 10, 11, 12, 13, 14, 15, 16, 17, 18, 19, \\ 20, 21, 22, 23, 24, 25, 26, 27, 28, 29, 30, 31, 32\end{tabular} \\ 
\hline
96\% 
& \begin{tabular}[c]{@{}c@{}}1, 2, 3, 4, 5, 6, 7, 8, 9, 10, 11, 12,   13, 14, 15, 16, 17, \\ 18, 19, 20, 21, 22, 23, 24, 25, 26, 27, 28, 29, 30, 31,  32\end{tabular} \\
\hline
\end{tabular}
\end{table}

%% file: tables/tab_eval_profiling_time.tex
\begin{table}[]
\caption{Number of runs and overall latency for each benchmark and each decomposition configuration. Note that the overall time does not refer to the entire time consumed for experiments.}
\label{tab:profiling_time}
\centering
\begin{tabular}{|l|r|r|}
\cline{1-3}
\multicolumn{1}{|c|}{\textbf{Benchmark}} 
& \multicolumn{1}{c|}{\textbf{\# Runs}} 
& \multicolumn{1}{c|}{\textbf{Latency for Multiple Runs (mins)}}
\\
\hline
ARC Easy
& 15
& 2
\\
\hline
ARC Challenge
& 20
& 2
\\
\hline
HellaSwag
& 2
& 3.5
\\
\hline
MMLU
& 2
& 8
\\
\hline
TruthfulQA
& 2
& 12
\\
\hline
WinoGrande
& 30
& 2
\\
\hline
\end{tabular}
\end{table}

%% file: sections/05_RelatedWorks.tex
\section{Related Works}
\label{sec:related_works}

\betterparagraph{Low-rank Adapters}
Low-rank decomposition has been used as an effective solution for fine-tuning LMs.
%
%
%
LoRA\cite{hu2022lora} is a recent low-rank adapter work that identifies the parameter adjustment during fine-tuning as intrinsically low-rank information and represents it using a low-rank-decomposed matrix. 
This approach accelerates fine-tuning processing while delivering high fine-tuning performance.
%
%
However, LoRA is a reparameterization approach, which does not change the final model size. 
LoRA keeps the original tensor size via merging the low-rank-decomposed matrix into the original weight, unlike our TKD approach directly compresses the model size.

%

\betterparagraph{Other Low-Rank Decomposition Usage in Deep Learning} Although low-rank decomposition has been actively explored as a model compression technique, there are few studies about their implication on recent large language models.
In \cite{ben-noach-goldberg-2020-compressing}, the authors decomposed the BERT model using SVD and perform feature distillation to recover accuracy.
However, the model scale and benchmark are not tailored for recent large language models and their tasks.
%
%
In \cite{hrinchuk-etal-2020-tensorized}, the authors proposed a way to decompose the input embedding layers using Tensor Train decomposition.
The authors in~\cite{phan2020stable} presented low-rank decomposition for compressing convolutional neural networks using Canonical Polyadic (CP) decomposition.
%
%
In~\cite{lin2018holistic}, the authors propose a methodology to decompose both the fully-connected layers and the convolutional layers and recover accuracy by retaining the decomposed model using knowledge transfer from the original model.
GroupReduce~\cite{NEURIPS2018_a2b8a85a} presents that language models with large vocabulary sizes have more than 90\% of the parameters in the embedding layer, and they can be compressed using low-rank decomposition.
%
%
TIE~\cite{8980328} proposes a computation-efficient inference scheme for DNNs decomposed using Tensor Train decomposition and a hardware accelerator architecture for the scheme.

\betterparagraph{Performance Characterization of LLMs} Because the large language model emerged recently, the computational performance characterization on LLM inference is not well-explored.
Megatron-LM analyzed the computational performance of LLM training targeting data centers~\cite{narayanan2021efficient}.
However, the work focused on the training workload without model compression techniques like low-rank decomposition.

%% file: sections/06_Conclusion.tex
\section{Conclusion and Future Work}
\label{sec:conclusion}

In this work, we explored the accuracy-efficiency trade-off of low-rank decomposition, which was not well understood previously.
By formalizing the decomposition design space, we showed that the space is huge, which makes it intractable to navigate.
To address the challenge, we perform characterization focusing on three axes (number of pruned ranks, the choice of decomposed layers, and decomposed tensors) identified based on our formulation and extract useful insights helpful for reducing the design space: we can prune the rank down to 1 with minimal accuracy impact, we should not decompose early layers, and we should not decompose adjacent layers.
Such insights can be adopted in future algorithm-level research to develop high-accuracy and efficient low-rank decomposition methods for LLMs.
In particular, recovering accuracy using fine-tuning after low-rank decomposition is a promising direction.
Our early investigation shows that we can recover the accuracy of a 15\% compressed model to that of a 9\% model within a single epoch of fine-tuning, which motivates future studies in this domain.

Our characterization code base will be open-sourced to facilitate such future studies.
In particular, unlike many other works, we include energy profiling, enabling a more thorough consideration of the accuracy-efficiency trade-off in future LLM algorithm research.

%% file: sections/07_Acknowledgement.tex
\section*{Acknowledgement}
We thank Prof. Sitao Huang and Prof. Yanning Shen for providing feedback on the paper.

%% file: main.bbl
\begin{thebibliography}{10}

\bibitem{openai2023gpt4}
OpenAI, ``Gpt-4 technical report,'' {\em arXiv preprint arXiv:2303.08774}, 2023.

\bibitem{huang2023realtime}
H.~Huang, F.~D.~L. Torre, C.~M. Fang, A.~Banburski-Fahey, J.~Amores, and J.~Lanier, ``Real-time animation generation and control on rigged models via large language models,'' in {\em NeurIPS 2023}, December 2023.
\newblock Spotlight Paper.

\bibitem{ni2023lever}
A.~Ni, S.~Iyer, D.~Radev, V.~Stoyanov, W.-t. Yih, S.~Wang, and X.~V. Lin, ``Lever: Learning to verify language-to-code generation with execution,'' in {\em International Conference on Machine Learning}, pp.~26106--26128, PMLR, 2023.

\bibitem{touvron2023llama}
H.~Touvron, L.~Martin, K.~Stone, P.~Albert, A.~Almahairi, Y.~Babaei, N.~Bashlykov, S.~Batra, P.~Bhargava, S.~Bhosale, {\em et~al.}, ``Llama 2: Open foundation and fine-tuned chat models,'' {\em arXiv preprint arXiv:2307.09288}, 2023.

\bibitem{he2016deep}
K.~He, X.~Zhang, S.~Ren, and J.~Sun, ``Deep residual learning for image recognition,'' in {\em Proceedings of the IEEE conference on computer vision and pattern recognition}, pp.~770--778, 2016.

\bibitem{vaswani2017attention}
A.~Vaswani, N.~Shazeer, N.~Parmar, J.~Uszkoreit, L.~Jones, A.~N. Gomez, {\L}.~Kaiser, and I.~Polosukhin, ``Attention is all you need,'' {\em Advances in neural information processing systems}, vol.~30, 2017.

\bibitem{frantar2023sparsegpt}
E.~Frantar and D.~Alistarh, ``Sparsegpt: Massive language models can be accurately pruned in one-shot,'' in {\em International Conference on Machine Learning}, pp.~10323--10337, PMLR, 2023.

\bibitem{kim2021bert}
S.~Kim, A.~Gholami, Z.~Yao, M.~W. Mahoney, and K.~Keutzer, ``I-bert: Integer-only bert quantization,'' in {\em International conference on machine learning}, pp.~5506--5518, PMLR, 2021.

\bibitem{yao2022zeroquant}
Z.~Yao, R.~Yazdani~Aminabadi, M.~Zhang, X.~Wu, C.~Li, and Y.~He, ``Zeroquant: Efficient and affordable post-training quantization for large-scale transformers,'' {\em Advances in Neural Information Processing Systems}, vol.~35, pp.~27168--27183, 2022.

\bibitem{dettmers2022gpt3}
T.~Dettmers, M.~Lewis, Y.~Belkada, and L.~Zettlemoyer, ``Gpt3. int8 (): 8-bit matrix multiplication for transformers at scale,'' {\em Advances in Neural Information Processing Systems}, vol.~35, pp.~30318--30332, 2022.

\bibitem{zare2018extension}
A.~Zare, A.~Ozdemir, M.~A. Iwen, and S.~Aviyente, ``Extension of pca to higher order data structures: An introduction to tensors, tensor decompositions, and tensor pca,'' {\em Proceedings of the IEEE}, vol.~106, no.~8, pp.~1341--1358, 2018.

\bibitem{kong2017low}
S.~Kong and C.~Fowlkes, ``Low-rank bilinear pooling for fine-grained classification,'' in {\em Proceedings of the IEEE conference on computer vision and pattern recognition}, pp.~365--374, 2017.

\bibitem{yu2017compressing}
X.~Yu, T.~Liu, X.~Wang, and D.~Tao, ``On compressing deep models by low rank and sparse decomposition,'' in {\em Proceedings of the IEEE conference on computer vision and pattern recognition}, pp.~7370--7379, 2017.

\bibitem{gusak2019automated}
J.~Gusak, M.~Kholiavchenko, E.~Ponomarev, L.~Markeeva, P.~Blagoveschensky, A.~Cichocki, and I.~Oseledets, ``Automated multi-stage compression of neural networks,'' in {\em Proceedings of the IEEE/CVF International Conference on Computer Vision Workshops}, pp.~0--0, 2019.

\bibitem{wang2023cuttlefish}
H.~Wang, S.~Agarwal, Y.~Tanaka, E.~Xing, D.~Papailiopoulos, {\em et~al.}, ``Cuttlefish: Low-rank model training without all the tuning,'' {\em Proceedings of Machine Learning and Systems}, vol.~5, 2023.

\bibitem{xu2023tensorgpt}
M.~Xu, Y.~L. Xu, and D.~P. Mandic, ``Tensorgpt: Efficient compression of the embedding layer in llms based on the tensor-train decomposition,'' {\em arXiv preprint arXiv:2307.00526}, 2023.

\bibitem{clark2018think}
P.~Clark, I.~Cowhey, O.~Etzioni, T.~Khot, A.~Sabharwal, C.~Schoenick, and O.~Tafjord, ``Think you have solved question answering? try arc, the ai2 reasoning challenge,'' {\em arXiv preprint arXiv:1803.05457}, 2018.

\bibitem{zellers2019hellaswag}
R.~Zellers, A.~Holtzman, Y.~Bisk, A.~Farhadi, and Y.~Choi, ``Hellaswag: Can a machine really finish your sentence?,'' in {\em Proceedings of the 57th Annual Meeting of the Association for Computational Linguistics}, 2019.

\bibitem{hendryckstest2021}
D.~Hendrycks, C.~Burns, S.~Basart, A.~Zou, M.~Mazeika, D.~Song, and J.~Steinhardt, ``Measuring massive multitask language understanding,'' {\em Proceedings of the International Conference on Learning Representations (ICLR)}, 2021.

\bibitem{lin-etal-2022-truthfulqa}
S.~Lin, J.~Hilton, and O.~Evans, ``{T}ruthful{QA}: Measuring how models mimic human falsehoods,'' in {\em Proceedings of the 60th Annual Meeting of the Association for Computational Linguistics (Volume 1: Long Papers)} (S.~Muresan, P.~Nakov, and A.~Villavicencio, eds.), (Dublin, Ireland), pp.~3214--3252, Association for Computational Linguistics, May 2022.

\bibitem{10.1145/3474381}
K.~Sakaguchi, R.~L. Bras, C.~Bhagavatula, and Y.~Choi, ``Winogrande: an adversarial winograd schema challenge at scale,'' {\em Commun. ACM}, vol.~64, p.~99–106, aug 2021.

\bibitem{google_agent_assist}
G.~Cloud, ``Agent assist.'' \url{https://cloud.google.com/agent-assist?hl=en}, 2024.

\bibitem{github_copilot}
Github, ``Github copilot.'' \url{https://github.com/features/copilot}, 2023.

\bibitem{microsoft_ai_assistant}
Microsoft, ``Announcing microsoft copilot, your everyday ai companion.'' \url{https://blogs.microsoft.com/blog/2023/09/21/announcing-microsoft-copilot-your-everyday-ai-companion/}, 2023.

\bibitem{kenton2019bert}
J.~D. M.-W.~C. Kenton and L.~K. Toutanova, ``Bert: Pre-training of deep bidirectional transformers for language understanding,'' in {\em Proceedings of naacL-HLT}, vol.~1, p.~2, 2019.

\bibitem{anil2023gemini}
R.~Anil, S.~Borgeaud, Y.~Wu, J.-B. Alayrac, J.~Yu, R.~Soricut, J.~Schalkwyk, A.~M. Dai, A.~Hauth, K.~Millican, {\em et~al.}, ``Gemini: A family of highly capable multimodal models,'' {\em arXiv preprint arXiv:2312.11805}, vol.~1, 2023.

\bibitem{kim2023full}
S.~Kim, C.~Hooper, T.~Wattanawong, M.~Kang, R.~Yan, H.~Genc, G.~Dinh, Q.~Huang, K.~Keutzer, M.~W. Mahoney, {\em et~al.}, ``Full stack optimization of transformer inference: a survey,'' {\em arXiv preprint arXiv:2302.14017}, 2023.

\bibitem{williams2009roofline}
S.~Williams, A.~Waterman, and D.~Patterson, ``Roofline: an insightful visual performance model for multicore architectures,'' {\em Communications of the ACM}, vol.~52, no.~4, pp.~65--76, 2009.

\bibitem{rajpurkar-etal-2016-squad}
P.~Rajpurkar, J.~Zhang, K.~Lopyrev, and P.~Liang, ``{SQ}u{AD}: 100,000+ questions for machine comprehension of text,'' in {\em Proceedings of the 2016 Conference on Empirical Methods in Natural Language Processing} (J.~Su, K.~Duh, and X.~Carreras, eds.), (Austin, Texas), pp.~2383--2392, Association for Computational Linguistics, Nov. 2016.

\bibitem{deng2009imagenet}
J.~Deng, W.~Dong, R.~Socher, L.-J. Li, K.~Li, and L.~Fei-Fei, ``Imagenet: A large-scale hierarchical image database,'' in {\em 2009 IEEE conference on computer vision and pattern recognition}, pp.~248--255, Ieee, 2009.

\bibitem{kao2023flat}
S.-C. Kao, S.~Subramanian, G.~Agrawal, A.~Yazdanbakhsh, and T.~Krishna, ``Flat: An optimized dataflow for mitigating attention bottlenecks,'' in {\em Proceedings of the 28th ACM International Conference on Architectural Support for Programming Languages and Operating Systems (ASPLOS), Volume 2}, pp.~295--310, 2023.

\bibitem{Harshman1970FoundationsOT}
R.~A. Harshman, ``Foundations of the parafac procedure: Models and conditions for an "explanatory" multi-model factor analysis,'' 1970.

\bibitem{GNNTensorTrain}
C.~Yin, D.~Zheng, I.~Nisa, C.~Faloutsos, G.~Karypis, and R.~Vuduc, ``Nimble gnn embedding with tensor-train decomposition,'' in {\em Proceedings of the 28th ACM SIGKDD Conference on Knowledge Discovery and Data Mining}, KDD '22, (New York, NY, USA), p.~2327–2335, Association for Computing Machinery, 2022.

\bibitem{nips2020tensortrain}
J.~Su, W.~Byeon, J.~Kossaifi, F.~Huang, J.~Kautz, and A.~Anandkumar, ``Convolutional tensor-train lstm for spatio-temporal learning,'' in {\em Advances in Neural Information Processing Systems} (H.~Larochelle, M.~Ranzato, R.~Hadsell, M.~Balcan, and H.~Lin, eds.), vol.~33, pp.~13714--13726, Curran Associates, Inc., 2020.

\bibitem{D-Tucker}
J.-G. Jang and U.~Kang, ``D-tucker: Fast and memory-efficient tucker decomposition for dense tensors,'' in {\em 2020 IEEE 36th International Conference on Data Engineering (ICDE)}, pp.~1850--1853, 2020.

\bibitem{doi:10.1137/06066518X}
V.~de~Silva and L.-H. Lim, ``Tensor rank and the ill-posedness of the best low-rank approximation problem,'' {\em SIAM Journal on Matrix Analysis and Applications}, vol.~30, no.~3, pp.~1084--1127, 2008.

\bibitem{Lebedev2014SpeedingupCN}
V.~Lebedev, Y.~Ganin, M.~Rakhuba, I.~Oseledets, and V.~S. Lempitsky, ``Speeding-up convolutional neural networks using fine-tuned cp-decomposition,'' {\em CoRR}, vol.~abs/1412.6553, 2014.

\bibitem{tucker-cp-comp}
E.~Hale and A.~Prater-Bennette, ``{Comparison of CP and Tucker tensor decomposition algorithms},'' in {\em Big Data III: Learning, Analytics, and Applications} (F.~Ahmad, P.~P. Markopoulos, and B.~Ouyang, eds.), vol.~11730, p.~117300D, International Society for Optics and Photonics, SPIE, 2021.

\bibitem{de2000multilinear}
L.~De~Lathauwer, B.~De~Moor, and J.~Vandewalle, ``A multilinear singular value decomposition,'' {\em SIAM journal on Matrix Analysis and Applications}, vol.~21, no.~4, pp.~1253--1278, 2000.

\bibitem{huggingface2024llmleaderboard}
H.~Face, ``Open llm leaderboard.'' \url{https://huggingface.co/open-llm-leaderboard}, 2024.

\bibitem{eval-harness}
L.~Gao, J.~Tow, B.~Abbasi, S.~Biderman, S.~Black, A.~DiPofi, C.~Foster, L.~Golding, J.~Hsu, A.~Le~Noac'h, H.~Li, K.~McDonell, N.~Muennighoff, C.~Ociepa, J.~Phang, L.~Reynolds, H.~Schoelkopf, A.~Skowron, L.~Sutawika, E.~Tang, A.~Thite, B.~Wang, K.~Wang, and A.~Zou, ``A framework for few-shot language model evaluation.'' https://zenodo.org/records/10256836, 12 2023.

\bibitem{hu2022lora}
E.~J. Hu, P.~Wallis, Z.~Allen-Zhu, Y.~Li, S.~Wang, L.~Wang, W.~Chen, {\em et~al.}, ``Lora: Low-rank adaptation of large language models,'' in {\em International Conference on Learning Representations}, 2022.

\bibitem{ben-noach-goldberg-2020-compressing}
M.~Ben~Noach and Y.~Goldberg, ``Compressing pre-trained language models by matrix decomposition,'' in {\em Proceedings of the 1st Conference of the Asia-Pacific Chapter of the Association for Computational Linguistics and the 10th International Joint Conference on Natural Language Processing} (K.-F. Wong, K.~Knight, and H.~Wu, eds.), (Suzhou, China), pp.~884--889, Association for Computational Linguistics, Dec. 2020.

\bibitem{hrinchuk-etal-2020-tensorized}
O.~Hrinchuk, V.~Khrulkov, L.~Mirvakhabova, E.~Orlova, and I.~Oseledets, ``Tensorized embedding layers,'' in {\em Findings of the Association for Computational Linguistics: EMNLP 2020} (T.~Cohn, Y.~He, and Y.~Liu, eds.), (Online), pp.~4847--4860, Association for Computational Linguistics, Nov. 2020.

\bibitem{phan2020stable}
A.-H. Phan, K.~Sobolev, K.~Sozykin, D.~Ermilov, J.~Gusak, P.~Tichavsk{\`y}, V.~Glukhov, I.~Oseledets, and A.~Cichocki, ``Stable low-rank tensor decomposition for compression of convolutional neural network,'' in {\em Computer Vision--ECCV 2020: 16th European Conference, Glasgow, UK, August 23--28, 2020, Proceedings, Part XXIX 16}, pp.~522--539, Springer, 2020.

\bibitem{lin2018holistic}
S.~Lin, R.~Ji, C.~Chen, D.~Tao, and J.~Luo, ``Holistic cnn compression via low-rank decomposition with knowledge transfer,'' {\em IEEE transactions on pattern analysis and machine intelligence}, vol.~41, no.~12, pp.~2889--2905, 2018.

\bibitem{NEURIPS2018_a2b8a85a}
P.~Chen, S.~Si, Y.~Li, C.~Chelba, and C.-J. Hsieh, ``Groupreduce: Block-wise low-rank approximation for neural language model shrinking,'' in {\em Advances in Neural Information Processing Systems} (S.~Bengio, H.~Wallach, H.~Larochelle, K.~Grauman, N.~Cesa-Bianchi, and R.~Garnett, eds.), vol.~31, Curran Associates, Inc., 2018.

\bibitem{8980328}
C.~Deng, F.~Sun, X.~Qian, J.~Lin, Z.~Wang, and B.~Yuan, ``Tie: Energy-efficient tensor train-based inference engine for deep neural network,'' in {\em 2019 ACM/IEEE 46th Annual International Symposium on Computer Architecture (ISCA)}, pp.~264--277, 2019.

\bibitem{narayanan2021efficient}
D.~Narayanan, M.~Shoeybi, J.~Casper, P.~LeGresley, M.~Patwary, V.~Korthikanti, D.~Vainbrand, P.~Kashinkunti, J.~Bernauer, B.~Catanzaro, {\em et~al.}, ``Efficient large-scale language model training on gpu clusters using megatron-lm,'' in {\em Proceedings of the International Conference for High Performance Computing, Networking, Storage and Analysis}, pp.~1--15, 2021.

\end{thebibliography}
